%% file: neurips_2026.tex
\documentclass{article}

  \usepackage[preprint]{neurips_2026}

\usepackage[utf8]{inputenc} % allow utf-8 input
\usepackage[T1]{fontenc}    % use 8-bit T1 fonts
\usepackage{hyperref}       % hyperlinks
\usepackage{url}            % simple URL typesetting
\usepackage{booktabs}       % professional-quality tables
\usepackage{amsfonts}       % blackboard math symbols
\usepackage{nicefrac}       % compact symbols for 1/2, etc.
\usepackage{microtype}      % microtypography
\usepackage{xcolor}         % colors
\input{math_commands.tex}

\usepackage[utf8]{inputenc} % allow utf-8 input
\usepackage[T1]{fontenc}    % use 8-bit T1 fonts
\usepackage{hyperref}       % hyperlinks
\usepackage{url}            % simple URL typesetting
\usepackage{booktabs}       % professional-quality tables
\usepackage{amsfonts}       % blackboard math symbols
\usepackage{nicefrac}       % compact symbols for 1/2, etc.
\usepackage{microtype}      % microtypography
\usepackage{xcolor}         % colors
\usepackage{tabularx}
\usepackage{amsmath}
\usepackage{bm}
 \usepackage{amsmath, amssymb}
\usepackage{amsthm}
\usepackage{graphicx}
\usepackage{subcaption}
\usepackage{array} % for >{\raggedright\arraybackslash}
\usepackage{algorithm}
\usepackage{algpseudocode}
\usepackage{caption}
\usepackage{tabularx}
\usepackage{colonequals}    
\usepackage{enumitem}
\usepackage{tikz}
\usepackage{wrapfig}
\usetikzlibrary{tikzmark, calc, decorations.pathreplacing}
\title{Nash: Neural Adaptive Shrinkage for Structured
High-Dimensional Regression}

\author{%
  William R.P. Denault  \\
Departments of Statistics and Human Genetics\\
University of Chicago\\
  \texttt{wdenault@uchicago.edu} \\
}

\theoremstyle{plain}
\newtheorem{theorem}{Theorem}[section]
\newtheorem{proposition}[theorem]{Proposition}
\newtheorem{lemma}[theorem]{Lemma}
\newtheorem{corollary}[theorem]{Corollary}
\theoremstyle{definition}

\theoremstyle{remark}
\newtheorem{remark}[theorem]{Remark}
\renewcommand{\url}[1]{}
\begin{document}

\maketitle

\begin{abstract}
 Sparse linear regression is a fundamental tool in data analysis. However, traditional approaches often fall short when covariates exhibit structure or arise from heterogeneous sources. In biomedical applications, covariates may stem from distinct modalities or be structured according to an underlying graph. We introduce \textit{Neural Adaptive Shrinkage} (Nash), a unified framework that integrates covariate-specific side information into sparse regression via neural networks. Nash adaptively modulates penalties on a per-covariate basis, learning to tailor regularization without cross-validation. We use a \textit{split variational empirical Bayes} algorithm that decouples prior learning from posterior inference, reducing the M-step from  $\mathcal{O}(p) $ neural-network passes per sweep to a single batched pass, a \textit{74 to 106× wall-clock speedup} over previously proposed coordinate ascent CAVI for 
p between $10^2$ and $10^4$. Experiments on real data demonstrate that Nash improves accuracy and adaptability over existing methods.
\end{abstract}

\section{Introduction }

Regularization techniques for linear models have been central in data analysis for decades~\citep{hoerl_ridge_1970,tibshirani_regression_1996,zou_regularization_2005}. They remain central in modern data analysis as they are competitive approaches when the sample size is limited, and the covariates are high-dimensional \citep{horvath_dna_2018}. Despite their popularity, these methods often fall short when dealing with heterogeneous covariates that exhibit structural properties, such as nominal, ordinal, spatial, or graphical data. Classical regularization methods like Lasso \citep{tibshirani_regression_1996} typically apply uniform penalties across all covariates, which can be suboptimal when diverse predictor types are present in the covariate matrix (e.g, different genetic modalities). Real-world problems often benefit from tailored regularization that leverages the covariate side information, such as geographical proximity \citep{devriendt_sparse_2021} or type biological measurements \citep{boulesteix_ipf-Lasso_2017}. On the other hand, the existing methods that leverage covariate side information  \citep{tibshirani_sparsity_2005,yuan_model_2006,yu_sparse_2016,boulesteix_ipf-Lasso_2017} are often limited by their application-specific nature and reliance on cumbersome cross-validation for hyperparameter selection \citep{tibshirani_sparsity_2005,yuan_model_2006}. %Often resulting in practitioners using ad hoc methods to select the hyperparameters e.g. the 1 standard error rule in Lasso (see \citet{hastie_elements_2009}, chapter 7) $\hat{\lambda}_{\text{1-SE}} = \max \left\{ \lambda : \text{CV-error}(\lambda) \leq \text{CV-error}(\hat{\lambda}_{\min}) + \text{SE}(\hat{\lambda}_{\min}) \right\}  $ often used in practice see \citep{bohlin_prediction_2016,horvath_dna_2018,haftorn_epic_2021}.

In this work, we introduce \textit{neural adaptive shrinkage} (Nash), a novel regression model framework that can leverage neural networks to automatically learn the form of the penalty and select the amount of regularization without using cross-validation or approximate methods. Hence, alleviating the limitations listed above. We fit Nash using a novel variational inference empirical Bayes (VEB) method called split VEB, originally developed for smoothing  over-dispersed Poisson count data in the absence of a design matrix~\citep{xie_empirical_2023}.

We show that adapting split VEB to high-dimensional Gaussian linear regression with 
covariate-dependent priors yield a coordinate ascent algorithm that reduces 
the cost of neural network prior learning from $O(p)$ backward passes per 
sweep to a single batched update results specific to the 
Gaussian linear setting and absent from \citet{xie_empirical_2023}.  When no side information is available, our approach corresponds to optimizing the lower bound of a recently proposed model by \citet{kim_flexible_2024} and has similar computation complexity $O\left((n +K)p\right)$, where $K$ is the number of components in the adaptive shrinkage prior \citet{stephens_false_2017} used by \citet{kim_flexible_2024}. However, our learning algorithm is much simpler than the one proposed by  \citet{kim_flexible_2024}  and allows easy integration of machine learning approaches for penalty learning (e.g., neural net, xgboost \citet{chen_xgboost_2016}). Hence, Nash is both an extremely efficient high-dimensional regression method when no side information is present and a very flexible alternative when side information is available. We demonstrate that Nash is a highly competitive framework through a comprehensive study on real data examples.

\section{Previous works and contribution}

\textbf{Previous works} have mostly focused on two main types of side information on the covariate. The first type corresponds to groups (e.g., DNA methylation vs genotype data \citep{boulesteix_ipf-Lasso_2017}) or hierarchical information on the covariate; these works include group Lasso \citep{yuan_model_2006}  and other of its variations 
\citep{gertheiss_sparse_2010,tutz_modelling_2017, oelker_uniform_2017} and the  IPF Lasso \citep{boulesteix_ipf-Lasso_2017}. Essentially, these methods extend classical regularized techniques for linear models by using different additive sub-penalties that depend on the group/hierarchy of the covariates. The second type of covariate side information leveraged in penalized regression is graphs \citep{tibshirani_sparsity_2005,tibshirani_solution_2011}. Spanning from simple  L0 graph filtering problems such as Fused Lasso \citep{tibshirani_sparsity_2005} to more complex graphical structures that can be handled by the GEN Lasso \citep{tibshirani_solution_2011} and more recent variations  \citep{yu_sparse_2016,devriendt_sparse_2021} that can fit a mix of the different penalties above within a single framework.

 \textbf{Our contribution.} While combining neural networks with linear regression is not new \citep{okoh_hybrid_2018,nalisnick_hybrid_2019,lemhadri_LassoNET_2021}, existing methods focus on hybrid models  \citep{okoh_hybrid_2018,nalisnick_hybrid_2019} or learning link functions \citep{lemhadri_LassoNET_2021} rather than learning the penalty itself. Our work differs substantially from the previous works listed above. To our knowledge, this is the first work to propose the use of a neural network to incorporate covariate side information when learning the penalty function in linear regression. Our work is much more assumption-lean compared to previous works, as Nash can leverage any side information that is processable by a neural net. Additionally, we propose a novel low-complexity variational approximation for empirical Bayes in multiple linear regression. The resulting learning algorithm is a simple and effective iterative procedure, akin to ADMM or proximal algorithms \citep{polson_proximal_2015}.

\section{Problem Definition}

\subsection{Target Model}\label{subsec:target_model}

The Nash model is defined as follows:
\begin{align}\label{eq:target_regression}
   \bm y | \bm X, \bm \beta, \sigma^2 &\sim N(\bm X \bm \beta, \sigma^2) \\ 
   %\beta_j &\sim N(b_j, \sigma^2_0) \\ 
  \beta_j &\sim g(\bm d_j, \bm \theta)
\end{align}

where $\bm y$ is a response vector of length $n$, $\bm X$ is an $n \times p$ matrix, where $p$ can be much larger than $n$ (i.e., $p \gg n$), and $\bm x_j$ is the $j^{th}$ column of $\bm X$. The term  $\sigma^2 > 0$ is a strictly positive variance parameter. The vector $\bm d_j$ corresponds to side information on column $j$. The function $g(\cdot, \cdot)$ belongs to a certain class of functions $\mathcal{G}$ and takes $\bm d_j$ (side information) as its first argument and $\bm \theta$ (parameters) as its second argument. For any tuple $(\bm d_j, \bm \theta)$, $g(\bm d_j, \bm \theta)$ defines a distribution with a density, denoted as $g(\beta_j; \bm d_j, \bm \theta)$ at the point $b_j$.

For simplicity, we assume that $\bm y$ is scaled, centered with unit variance, and similarly that each column of $\bm X$. Note that we do not model the intercept in \eqref{eq:target_regression}, as centering $\bm y$ and $\bm X$ prior to model fitting accounts for it, and it is straightforward to recover the effect for the unscaled $\bm X$ \citep{chipman_practical_2001}.

We assume that for each predictor $\bm x_j$ in $\bm X$, we observe some side information $\bm d_j$. We intentionally remain vague on the form of the side information $\bm d_j$, with the only constraint being that $\bm d_1, \ldots, \bm d_p$ can be processed by a neural network (e.g., images, tokens, graph matrices). For ease of presentation, we assume that we can store $\bm d_1, \ldots, \bm d_p$ in a matrix $\bm D$ of size $p \times k$. The case without any side information can be recovered by setting $\bm d_1 = \bm d_2 = \ldots = \bm d_p$, i.e., constant side information.

Assuming that $\sigma^2$  is known , solving \eqref{eq:target_regression} in an Empirical Bayes (EB) fashion involves the following steps:

\begin{enumerate}
    \item[\textcolor{red}{1}] Learning the parameter $\bm \theta$ of the function $g(\cdot, \cdot) \in \mathcal{G}$ via maximum marginal likelihood $\mathcal{L}({\bm \theta})$
    \begin{align}
        \hat{\bm \theta} =& \argmax_{\bm \theta} \mathcal{L}({\bm \theta}) \\
        =& \argmax_{\bm \theta} \int   p(\bm y | \bm X, \bm \beta , \sigma^2)   g(\bm \beta ; \bm D, \theta) \, d \bm \beta\nonumber
    \end{align}
    
    \item[\textcolor{blue}{2}] Compute the posterior distribution
    \begin{align}
        p_{\text{post}}(\bm \beta ) = p(\bm \beta |\bm y, \bm X, \bm D, \sigma^2) \\
        \propto p(\bm y | \bm X, \bm \beta, \sigma^2)   g( \bm \beta; \bm D , \hat{\bm \theta})
    \end{align}
\end{enumerate}

Both steps are computationally intractable in general, even when $g$
does not depend on side information \citep{wang_empirical_2021,
kim_flexible_2024}, except in some very special cases. Our work presents
an additional difficulty: when $g$ is covariate-dependent, the
coordinate-ascent algorithm of \citet{kim_flexible_2024} for VEB lead to  couples the
prior update to every coordinate, requiring $p$ neural-network backward
passes per coordinate ascent sweep for $\bm \beta$ if applied to \ref{eq:target_regression}, which is intractable for
large $p$. 
Briefly, we circumvent these problems by  introducing a latent Gaussian that serves as a computational
device to decouple prior learning from posterior inference, resulting in a much more efficient learning algorithm that requires only a single neural-network update per coordinate ascent sweep.

\subsection{Split Variational inference}

To fit the target model \eqref{eq:target_regression} efficiently, we
introduce an augmented model that adds a latent layer $\bm b$ between
the likelihood and the prior. This auxiliary variable acts purely as a
computational devices as  it enables a conjugate Gaussian bottleneck that
decouples prior learning from posterior inference. The result is a
coordinate-ascent algorithm requiring only a single neural network
update per sweep, compared to the $p$ updates that direct CAVI on
\eqref{eq:target_regression} would require.

\begin{equation}\label{eq:augmented_model}
\begin{array}{cc}
\textbf{Target model} & \textbf{Augmented model (Nash)} \\[6pt]
\begin{aligned}
  \bm y \mid \bm X, \bm \beta, \sigma^2
    &\sim \mathcal{N}(\bm X \bm \beta,\, \sigma^2) \\
  \beta_j
    &\sim g(\bm d_j,\, \bm \theta)
\end{aligned}
&
\begin{aligned}
  \bm y \mid \bm X, \bm \beta, \sigma^2
    &\sim \mathcal{N}(\bm X \bm \beta,\, \sigma^2) \\
  {\color{teal}\beta_j \mid b_j}
    &{\color{teal}\sim \mathcal{N}(b_j,\, \sigma^2_0)} \\
  b_j
    &\sim g(\bm d_j,\, \bm \theta)
\end{aligned}
\end{array}
\end{equation}

Given that we aim to fit the Nash  model \eqref{eq:augmented_model} in a tractable and efficient way, we use a standard meanfield approximation of the posterior, i.e.,   $ q(\bm \beta, \bm b ) = \prod_j^P q_{\beta_{j}}(\beta_j) q_{b_{j}}(b_j) $. This factorization leads to an the Evidence lower bound (ELBO, \citep{blei_variational_2017}) of the form

\begin{equation}
\scalebox{0.85}{$\displaystyle
\begin{array}{c @{\,} c}
\textbf{ ELBO: Target model} & \textbf{ ELBO : Augmented model (Nash)} \\[8pt]
\begin{aligned}
  \mathcal{F}_{\text{target}} = \,
  &\sum_i \mathbb{E}\!\left[\log p(y_i \mid \bm x_i,
    \bm\beta, \sigma^2)\right]+\tikzmark{tA} \\[10pt]
  \,&\sum_j \mathbb{E}\!\left[\log
    \frac{g(\beta_j;\,\bm d_j,\bm\theta)}
         {q_{\beta_j}(\beta_j)}\right]\tikzmark{tB}
\end{aligned}
&
\begin{aligned}
  \mathcal{F}_{\text{Nash}} = \,
  &\sum_i \mathbb{E}\!\left[\log p(y_i \mid \bm x_i,
    \bm\beta, \sigma^2)\right]+\tikzmark{nA} \\[10pt]
  \,&{\color{teal}
    \sum_j \mathbb{E}\!\left[\log
      \frac{p(\beta_j \mid b_j,\,\sigma_0^2)}
           {q_{\beta_j}(\beta_j)}\right]}+\tikzmark{nB} \\[10pt]
  \,&\sum_j \mathbb{E}\!\left[\log
      \frac{g(b_j;\,\bm d_j,\bm\theta)}
           {q_{b_j}(b_j)}\right]\tikzmark{nC}
\end{aligned}
\end{array}
\begin{tikzpicture}[overlay, remember picture,
                    every node/.style={font=\small}]

  \coordinate (TA) at (pic cs:tA);
  \coordinate (TB) at (pic cs:tB);
  \coordinate (NA) at (pic cs:nA);
  \coordinate (NB) at (pic cs:nB);
  \coordinate (NC) at (pic cs:nC);

  % Both rails on RIGHT side of their respective columns
  % Target rail: just right of target equations
  \coordinate (Lrail) at ([xshift=-165pt]TA);
  % Nash rail: just right of Nash equations
  \coordinate (Rrail) at ([xshift=15pt]NB);

  \coordinate (LRrail) at ([xshift=-120pt]NB);
  % Target: right brace (no mirror)
\draw[black, thick, decorate,
        decoration={brace, amplitude=7pt, mirror}]
    ([yshift=10pt]Lrail |- TA) -- ([yshift=-10pt]Lrail |- TB)
    node[midway, left=12pt, black, align=right]
      {steps \textcolor{red}{1} \& \textcolor{blue}{2}\\coupled};

  % Nash: red brace rows 1+2
  \draw[red, thick, decorate,
        decoration={brace, amplitude=7pt}]
    ([yshift=10pt]Rrail |- NA) -- ([yshift=-10pt]Rrail |- NB)
  node[midway, right=12pt, red, align=left, yshift=5pt]
  {step 1:\\update $q_{\bm\beta}$};

  % Nash: blue brace rows 2+3
  \draw[blue, thick, decorate,
        decoration={brace, amplitude=7pt, mirror}]
    ([yshift=10pt]LRrail |- NB) -- ([yshift=-10pt]LRrail |- NC)
       node[midway, left=12pt, blue, align=right]
      {step 2:\\update\\ $g,\,q_{\bm b}$};

\end{tikzpicture}
$}
\end{equation}\label{eq:split_explained}
Where the ELBO for the target model is taken using the following candidate posterior $
    q(\bm \beta  ) = \prod_j^P q_{\beta_{j}}(\beta_j)  $ . The main idea behind split VEB is to decouple the prior/penalty learning step (step 1) from the posterior computation step (step 2) described \ref{subsec:target_model}. We highlight in blue in equation \ref{eq:split_explained}, how the introduction of a Gaussian latent state splits the objective function into two parts that are decoupled, the two key steps of VEB. At a high level, split VEB allows deriving a coordinate ascent that essentially iterates between solving two simple problems similar to  optimization  techniques (e.g., ADMM or proximal algorithms \citet{polson_proximal_2015}).

\paragraph{High-Level Coordinate Ascent Update for Nash}

Let $\bar{\beta}_j = \mathbb{E}_q(\beta_j)$ denote the expected value of $\beta_j$ with respect to $q$, and $\bar{b}_j = \mathbb{E}_q(b_j)$ denote the expected value of $b_j$. We define $ \bm{\bar r} = \bm{y} - \bm{X} \bm{\bar \beta}$ as the vector of expected residuals with respect to $q$. Let $\bm{X}_{-j}$ be the design matrix excluding the $j^{th}$ column, and $q_{-j}$ denote all factors $q_{j'}$ except factor $j$. The expected residuals accounting for the linear effect of all variables other than $j$ are given by:

\begin{align}\label{eq:partial_residu}
    \bar{\bm{r}}_j = \bm{y} - \bm{X}_{-j} \bm{\bar\beta}_{-j} = \bm{y} - \sum_{j' \neq j} \bm{x}_{j'} \bar\beta_{j'}
\end{align}

\begin{enumerate}
    \item \textbf{Update for $q^*_{\beta_j}$:}  
    the coordinate ascent update $q^*_{\beta_j} = \arg \max_{q_{\beta_j}} F_{\text{Nash}}(g, q, \sigma^2, g, q, \sigma_0^2)$ is obtained by computing the posterior using
    $$
    p(\bar{\bm{r}}_j | \bm{x}_j, \beta_j, \sigma) p(\beta_j | \bar{b}_j, \sigma_0^2)
    $$
    This is a simple posterior computation due to conjugacy and has a closed form that only requires computing the ordinary least squares (OLS)  regression of $\bm{x}_j$ on $\bar{\bm{r}}_j$ (see remark \ref{sec:beta}).
    
    \item \textbf{Update for $(g^*, q_{\bm{b}}^*)$:}  
    The coordinate ascent update
    $$
    (g^*, q_{\bm{b}}^*) = \arg \max_{g, q_{\bm{b}}} F(q_{\bm{\beta}}, q_{\bm{b}}, g; \sigma^2)_{\text{Nash}}
    $$
    is obtained by fitting a neural net with the following objective function:
    \begin{align}
    \hat{\bm{\theta}} = & \arg \max_{\bm{\theta}} \mathcal{L}(\bm{\theta}) \label{eq:loss_g}\\
    = & \arg \max_{\bm{\theta}} \prod_{j=1}^p \int \mathcal{N}(\bar{\beta}_j; b_j, \sigma_0^2) \, g(b_j; \bm{d}_j, \bm{\theta}) \, db_j \nonumber
    \end{align}
    Then, by computing the posterior  of each $b_j$,
    $$
    p(b_j | \bar{\beta}_j, \bm{d}_j, \sigma_0^2) \propto \mathcal{N}(\bar{\beta}_j; b_j, \sigma_0^2) \, g(b_j; \bm{d}_j, \hat{\bm{\theta}})
    $$
  which is also a simple posterior computation.
    
    \item \textbf{Update for $\sigma^2$}
    \begin{enumerate}
        \item $(\sigma^2)^* = \arg \max_{\sigma^2} F(q_{\bm{\beta}}, q_{\bm{b}}, g, \sigma^2, \sigma_0^2)_{\text{Nash}}$
        
    \end{enumerate}
\end{enumerate}

The first step is a direct consequence of the work by \citet{kim_flexible_2024} (see Appendix for more details), the second step results from our splitting approach, and the last step is a standard coordinate ascent variational inference (CAVI) step. We provide the closed-form formulas for both $\sigma^2$ and $\sigma_0^2$  in the supplementary section \ref{section:detailed}. We describe the overall learning algorithm in the Appendix, Algorithm \ref{alg:Nash_CAVI}.

\paragraph{Choice of $\mathcal{G}$}
For clarity, suppose that $g(\cdot, \cdot)$ belong to a family of distributions $\mathcal{G}$ that have the following form:

\begin{align}\label{eq:prior_form}
    g(\bm{d}_j, \bm{\theta}) & = \sum_{m=0}^M \pi_m(\bm{d}_j, \bm{\theta}) g_m  \\
    \bm{\pi}(\bm{d}_j, \bm{\theta}) & = (\pi_0(\bm{d}_j, \bm{\theta}), \ldots, \pi_M(\bm{d}_j, \bm{\theta}))
\end{align}

where $g_m $ are fixed known distributions (e.g., $g_0 = \delta_0$ and $g_m = \mathcal{N}(0, \sigma_m^2)$ with $\sigma_m^2 < \sigma_{m+1}^2$ for all $m > 0$). Then $\bm{\pi}(., \bm{\theta})$ is a neural network that takes side information $\bm d_j$ as input and outputs a vector of probabilities $(\pi_0(\bm{d}_j, \bm{\theta}), \ldots, \pi_M(\bm{d}_j, \bm{\theta}))$ that sum to 1 (e.g., using a softmax function). Under this model, the loss for $\bm{\theta}$ has the following simple form:

\begin{align}
    \hat{\bm{\theta}} = \arg\max_{\bm{\theta}} \sum_{j=1}^P \log \sum_{m=0}^M \pi_m(\bm{d}_j, \bm{\theta}) L_{jm}
\end{align}

where $L_{jm}$ is defined as, $ 
    L_{jm} = \int p(\bar{\beta}_j | b_j) g_m(b_j) \, db_j $, the marginal likelihood of $\bar{\beta}_j$ under mixture component $m$. For Gaussian mixture components $g_m = \mathcal{N}(0, \sigma_m^2)$, we have that $
    L_{jm} = \mathcal{N}(\bar{\beta}_j; 0, \sigma_0^2 + \sigma_m^2) $. These integrals often cannot be computed analytically for other priors and error models. However, $(L_{jm})$ are simple one-dimensional integrals that are fast to approximate.   More generally, $g(\cdot, \cdot)$ can be any probabilistic model (e.g., Gaussian Process, but potentially more complex models) for which the loss in \ref{eq:loss_g} can be evaluated and the posterior $p(b_j|\bar{\beta}_j, \bm{d}_i, \sigma_0^2) \propto \mathcal{N}(\bar{\beta}_j; b_j, \sigma_0^2) g(b_i; \bm{d}_i, \hat{\bm{\theta}})$ can be computed. For computational efficiency, it is useful that both of these steps can be evaluated via closed-form formulas. Note that when $g$ does not depend on the covariate, then step 2) in the coordinate ascent described above corresponds to an empirical Bayes normal mean problem (EBNM; see \citet{willwerscheid_ebnm_2024} for an overview).% In this case, fitting $g$ as in \ref{eq:prior_form} with fixed $g_m(\cdot)$ distributions corresponds to estimating the mixture proportions for the different mixture components $g_m(\cdot)$. Using fixed distributions is particularly practical as it allows efficient estimation of the mixture components $(\pi_0, \ldots, \pi_M)$ via sequential quadratic programming, which is often achieved in sub-linear time (in terms of $p$), see \citet{kim_fast_2019}. 

%\begin{figure}
   % \centering
    %\includegraphics[width= \linewidth]{figures/figure1b_for_paper.pdf}
    %\caption{%Upper panel: Adaptation of Figure 1 from \citet{kim_flexible_2024}, showcasing that posterior mean shrinkage operators (left panel) for different choices of $\sigma_1^2, \ldots ,\sigma_M^2$ and $\pi_0, \ldots, \pi_M$ can mimic the shrinkage operators from some commonly used penalties (right-hand panel). Bottom panel left: 
   % Illustration of how Nash can mimic fused Lasso penalty when used with a %graph neural net prior-based. The left panel presents the induced prior %density from equation \ref{eq:Nash-fused}, allowing Nash to mimic the %fused Lasso penalty ( using $s_1 = 0.45$ and $s_2 = 0.15$). Right panel,  %penalty surface of the fused Lasso (i.e., $|b_1| +|b_2|+|b_1-b_2|$). }
%    \label{fig1}
%\end{figure} 
% \input{additional_details_updates}

 \vspace{1em}
\begin{remark}[Efficiency and adaptive dampening of the coordinate ascent]\label{sec:beta}
The update for $\beta_j$ at iteration $t+1$ takes the form
$\bar{\beta}_j^{t+1} = \omega_t \hat{\beta}^{t+1}_{j,\mathrm{MLE}} + (1-\omega_t)\bar{b}_j^t$,
where $\omega_t = (n-1)\sigma_{0t}^2/(\sigma_t^2 + (n-1)\sigma_{0t}^2)$ and
$\hat{\beta}_{j,\mathrm{MLE}} = \bm{x}_j^\intercal \bar{\bm{r}}_j/(n-1)$ is the OLS sufficient statistic.
Each update therefore, reduces to a single dot product, giving an overall
coordinate ascent complexity of $O(np)$, substantially cheaper than the
$O(n^2p+n^3)$ or $O(np^2+p^3)$ cost of the exact Bayesian ridge posterior
\citep{hoerl_ridge_1970}.
%Because $\sigma_t^2$ and $\sigma_{0t}^2$ are updated alongside $\beta$ and $g$ (steps 3a--3b), the weight $\omega_t$ is continuously re-estimated from the data: it governs how much novel evidence enters each update of $g$ and $q_{\bm{b}}$ versus how much the previous iterate is retained, acting as an automatic, data-driven dampening parameter.
\end{remark}

\paragraph{Comparison and Connection with mr.ash}
Nash is closely related to mr.ash \citep{kim_flexible_2024}, but differs in two
key respects. First, mr.ash cannot incorporate side information. Second, and more
technically, mr.ash's standard CAVI couples prior and posterior learning, requiring
$p$ updates of $g$ per sweep, or equivalently, $p$ neural-network backward passes
when side information is present. Split VEB decouples these two steps, reducing the
cost to a single update of $g$ per sweep regardless of $p$, which is what makes
neural-network-based priors tractable (see section \ref{subsec:mstep} for runtime comparison). Formally, when no side information is
provided, fitting Nash with split VEB optimizes a lower bound of mr.ash's ELBO
under the same prior family $\mathcal{G}$ (see Appendix~\ref{sec:connection_mrash}
and Algorithm~2 for a detailed comparison).

\section{Connection to Penalized Linear Regression and Beyond}\label{sec:plrandbeyond}

\citet{kim_flexible_2024} showed that mr.ash (and therefore Nash) can be viewed as a penalized linear regression (PLR) problem. When using an adaptive shrinkage prior (ash, \citep{stephens_false_2017}) of the form $g = \pi_0 \delta_0 + \sum_{m=1}^M \pi_m N(0, \sigma_m^2)$, different choices of $(\pi_0, \ldots, \pi_M)$ roughly correspond  to different penalties such as Ridge regression \citep{hoerl_ridge_1970}, L0Learn \citep{hazimeh_l0learn_2023}, Lasso \citep{tibshirani_regression_1996}, Elastic Net \citep{zou_regularization_2005}, the smoothly clipped absolute deviation (SCAD) penalty, and the minimax concave penalty (MCP) \citep{breheny_coordinate_2011}. The advantage of mr.ash and Nash is that the user doesn't need to specify the penalty, as the model learns the mixture $(\hat{\pi}_0, \ldots, \hat{\pi}_M)$ that best fits the data via EB. \citep{kim_flexible_2024} proposed the concept of a shrinkage operator to properly establish the connection between EB multiple linear regression and PLR. We further build on this idea by suggesting that some parameterizations of Nash (detailed below) can be viewed as extensions to previous PLR methods with side information.

\paragraph{Group-Based and Hierarchical Penalty}

Several approaches have been developed to modulate the penalty based on groups or hierarchical structures of the data. Examples include the Group Lasso \citep{yuan_model_2006}, which uses a penalty of the form $\lambda_1 \|\bm{b}\|_1 + \lambda_2 \sum_{k \in \mathcal{K}} \sqrt{d_k} \|\bm{b}_k\|_2$, and the IPF-Lasso \citet{boulesteix_ipf-Lasso_2017} with a penalty of the form $\lambda \sum_{k \in \mathcal{K}} \sum_{j \in k} \omega_k |b_j|$, where $\mathcal{K}$ corresponds to the different groups or clusters. These cases are easily handled by Nash, as they simply correspond to fitting an ash prior per group/cluster/category. This is achieved using a prior of the form $g_k(\cdot) = \pi_{0k} \delta_0 + \sum_{m=1}^M \pi_{mk} N(0, \sigma_m^2)$ for each $k$. In other words the side information $\bm{d}_j$ for the covariate $\bm{x}_j$ is a vector of length $\mathcal{K}$ with binary entries, where the $k^{\text{th}}$ entry of $\bm{d}_j$ is set to 1 if covariate $j$ belongs to group $k$. Thus, the model $\pi: \bm{d}_j \rightarrow (\pi_{0}(\bm{d}_j), \ldots, \pi_M(\bm{d}_j))$ is a multinomial regression that is straightforward to fit using standard machine learning routines. Unlike the Group Lasso or the IPF-Lasso, Nash can naturally fit different penalty types to different groups ( e.g., fitting an $L_1$ like penalty on group 1 and fitting an $L_2$ like penalty on group 2).

    \begin{wrapfigure}{r}{0.55\textwidth}
  \centering
  \vspace{-1em}
  \includegraphics[width= \linewidth]{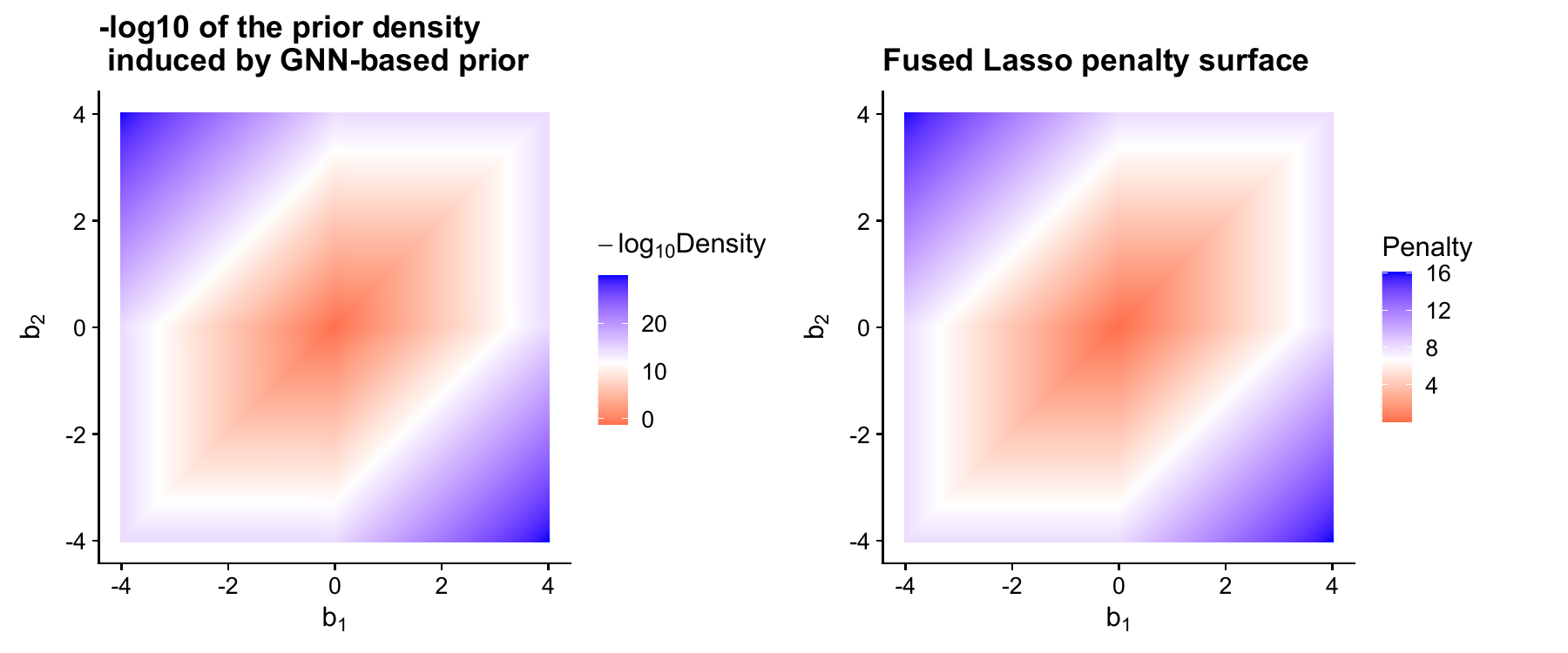}
   \caption{%Upper panel: Adaptation of Figure 1 from \citet{kim_flexible_2024}, showcasing that posterior mean shrinkage operators (left panel) for different choices of $\sigma_1^2, \ldots ,\sigma_M^2$ and $\pi_0, \ldots, \pi_M$ can mimic the shrinkage operators from some commonly used penalties (right-hand panel). Bottom panel left: 
    Illustration of how Nash can mimic the fused Lasso penalty when used with a graph neural net prior-based. The left panel presents the induced prior density from equation \ref{eq:Nash-fused}, allowing Nash to mimic the fused Lasso penalty ( using $s_1 = 0.45$ and $s_2 = 0.15$). Right panel,  penalty surface of the fused Lasso (i.e., $|b_1| +|b_2|+|b_1-b_2|$). }
    \label{fig1}
  \vspace{-2em}
\end{wrapfigure}
\paragraph{Fused Lasso and Graph-Based Penalty}\label{subsec:fused_Lasso}

The Fused Lasso \citep{tibshirani_sparsity_2005}  aims to balance sparsity and smoothness in covariates effects using a penalty of the form  $
\sum_{j=1}^p |b_j| \leq s_1 
\quad \text{and} \quad 
\sum_{j=2}^p |b_j - b_{j-1}| \leq s_2
$ . Bayesian versions \citet{casella_penalized_2010,betancourt_bayesian_2017} have been proposed. We extend these with graph neural networks (GNNs) to handle more complex dependencies. Classical Bayesian Fused Lasso \citet{casella_penalized_2010} can be reframed using a trivial graphical neural network (GNN) \citep{kipf_semi-supervised_2017}. Here, $\bm{d}_j = \bm{d}_j^{t+1}$ is the graph (a line in the Fused Lasso case) with nodes storing $\beta_{j,MLE}^{t+1}$ and $\bar{b}_j^t$. As the model converges, $\bar{b}_{j+1}^t \approx \bar{b}_{j+1}^{t+1}$, aligning with classic formulations. A generalized  GNN-EB Fused Lasso can be formulated as:
\begin{equation}\label{eq:Nash-fused}
    g_{\text{fused}}(\bm{d}_j) =  z L(0, s_1)L(l(\bm{d}_j), s_2)L(r(\bm{d}_j), s_2)
\end{equation}
Here, $L(\mu, s_0)$ is a Laplace distribution centered at $\mu$ with scale $s_0$, and $z$ is a normalization constant. Functions $r(\bm{d}_j) = \bar{b}^t_{j-1}$ and $l(\bm{d}_j) = \bar{b}^t_{j+1}$ are trivial GNNs, allowing different strengths for previous and next values. Posterior moments for $b_j$ can be approximated via Gauss-Hermite quadrature. Hyperparameters $( s_1, s_2)$ are learned by maximizing the marginal log likelihood. For an arbitrary graph, model \ref{eq:Nash-fused} becomes computationally challenging as computing the posterior under a product of $k>3$ Laplace distributions, as it quickly becomes computationally demanding to approximate. We propose \textbf{a restricted GNN- EB Fused Lasso} :
\begin{equation}\label{eq:g_fused}
    g_{\text{fused}}(\bm{d}_j) =z  L(0, s_1)L(v_1(\bm{d}_j), s_2(\bm{d}_j)) 
\end{equation}
Here, $v_1 (\bm{d}_j),s_2(\bm{d}_j)$ is the output of a GNN output controlling $ b_j$'s smoothness with respect to the graph structure. This simplifies normalization computation and integral approximation as it only uses two Laplace distributions. % Note that different variations of the Fused Lasso have been proposed, such as the Sparse Regression Incorporating Graphical Structure Among Predictors (SRIG)  \citet{yu_sparse_2016} or the Graph-Guided Fused Lasso (GGFL) \citet{chen_graph-structured_2010}. The  SRIG and GGFL penalties can also be mimicked by adapting the prior \ref{eq:g_fused} using Normal instead of Laplace.

\paragraph{Beyond Regularization}

We also provide an implementation of Nash that uses penalties based on Mixture Density Networks (MDN) \citep{bishop_mixture_1994} and Graph Mixture Density Networks (GMDN) \citep{errica_graph_2021}. The formulation is as follows:

\begin{align}\label{Nashmdn}
    g(j,\bm{\theta}) = \pi_0(\bm{d}_j) \delta_0 + \sum_k \pi_k(\bm{d}_j) N(\mu_k(\bm{d}_j), \sigma_k^2(\bm{d}_j))
\end{align} 

Here, $(\pi_k(\bm{d}_j)), (\mu_k(\bm{d}_j)), (\sigma_k^2(\bm{d}_j))$ are the outputs of the (graph) MDN, as described in \citet{bishop_mixture_1994} and in \citet{errica_graph_2021}. These parameterizations allow the model to actually push the values of $b_j$ away from 0. Enabling Nash to be used as a "self-supervised" biased regression where the bias (here $\mu_k(\bm{d}_j)$) is automatically learned from the data.

\input{synthethic_experiment}
\section{Discussion}

%IN APPENDIX REFERNCE AT CEBMF AND cebnm solvers

We proposed Nash, a novel high-dimensional regression framework that integrates covariate-specific side information into neural network-based estimation. Nash adaptively learns structured penalties nonparametrically, enabling flexible regularization without cross-validation. Our method generalizes and extends existing approaches that incorporate side information, providing a unified, more expressive framework. A central algorithmic contribution is the adaptation of split VEB to decouple prior learning from posterior inference, yielding a single batched M-step that is orders of magnitude faster than allowed by previous coordinate ascent (Table~\ref{tab:runtime}), while supporting far richer prior families  than \citep{kim_flexible_2024}
and naturally amenable to GPU acceleration. 

To our knowledge, Nash is the first \textbf{linear regression} framework 
to parameterize the \textbf{prior distribution} over regression effects 
as an arbitrary function of covariate side information. Nash is highly 
\textbf{modular}: the user needs only supply a solver for the cEBNM 
problem under their chosen prior family, a class that can be very 
large, encompassing neural networks, gradient boosting, Gaussian 
processes, or any structured probabilistic model for which the marginal 
likelihood and posterior moments are tractable. This enables automatic, 
data-driven regularization across diverse structures such as groups, 
time, and graphs, without requiring the penalty form to be specified 
in advance.
  %This algorithm naturally connects to and simplifies a recently proposed variational Empirical Bayes approach\citep{kim_flexible_2024}, while supporting far richer prior families and being highly \textbf{ modular}.  
\textbf{Limitations:}  Nash's current implementation is restricted to Gaussian likelihoods. Additionally, to maintain computational efficiency,  Nash requires a tractable cEBNM solver  \cite{denault_covariate-moderated_2025} for the chosen prior family, which requires either a closed form or fast  approximation of the marginal log-likelihood, which may not always be available.
\bibliographystyle{abbrvnat}
\bibliography{library}
\newpage 
 
%%%%%%%%%%%%%%%%%%%%%%%%%%%%%%%%%%%%%%%%%%%%%%%%%%%%%%%%%%%
 
%%%%%%%%%%%%%%%%%%%%%%%%%%%%%%%%%%%%%%%%%%%%%%%%%%%%%%%%%%%%
 \input{appendix}

\input{experiments}

\end{document}

%% file: math_commands.tex
\usepackage{amsmath,amsfonts,bm}

\def\eqref#1{equation~\ref{#1}}
\def\1{\bm{1}}

\DeclareMathAlphabet{\mathsfit}{\encodingdefault}{\sfdefault}{m}{sl}
\SetMathAlphabet{\mathsfit}{bold}{\encodingdefault}{\sfdefault}{bx}{n}

\DeclareMathOperator*{\argmax}{arg\,max}

%% file: synthethic_experiment.tex
\section{Numerical Experiment} 
 \begin{figure}
    \centering
    \includegraphics[width=1\linewidth]{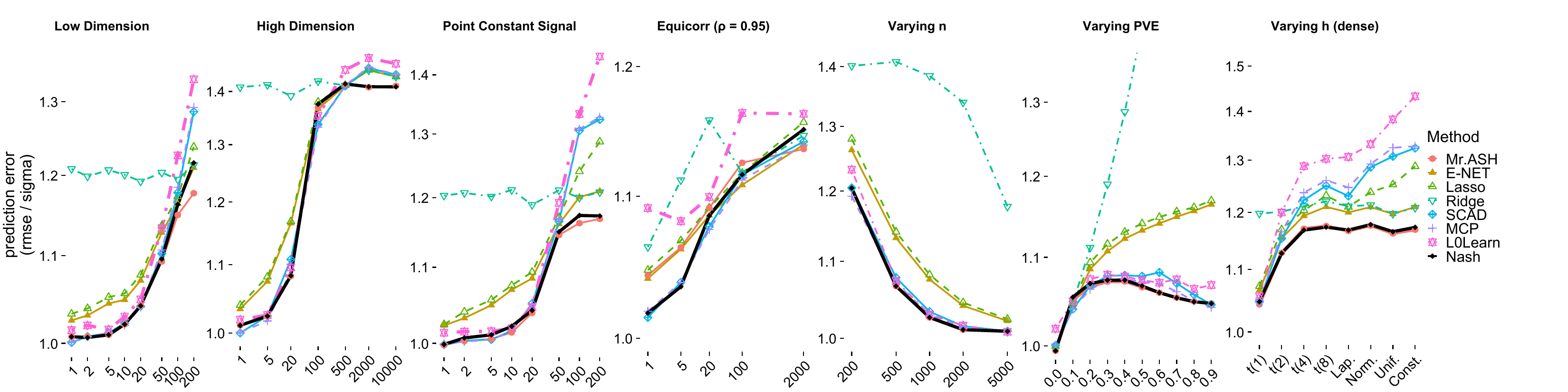}
\caption{Prediction accuracy of Nash against competing penalized regression
methods (Mr.ASH, Elastic Net, Lasso, Ridge, SCAD, MCP, L0Learn) across
simulated datasets, measured by scaled RMSE (lower is better). Additional results are presented in Supplementary Figures \ref{fig:placeholder1} }
    \label{fig:placeholder}
    \vspace{-1em}
\end{figure}

\subsection{Comparison with baselines}

We assessed the prediction accuracy of Nash against seven  competing methods
benchmarked in \citet{kim_flexible_2024}. To ensure a fair comparison, we ran Nash on
the identical simulated datasets using the same random seeds as \citet{kim_flexible_2024}, and
directly reuse their published results for all competing methods, eliminating variation due to sampling.
\begin{wrapfigure}{r}{0.6\textwidth}
  \centering 
  \includegraphics[width=0.6\textwidth]{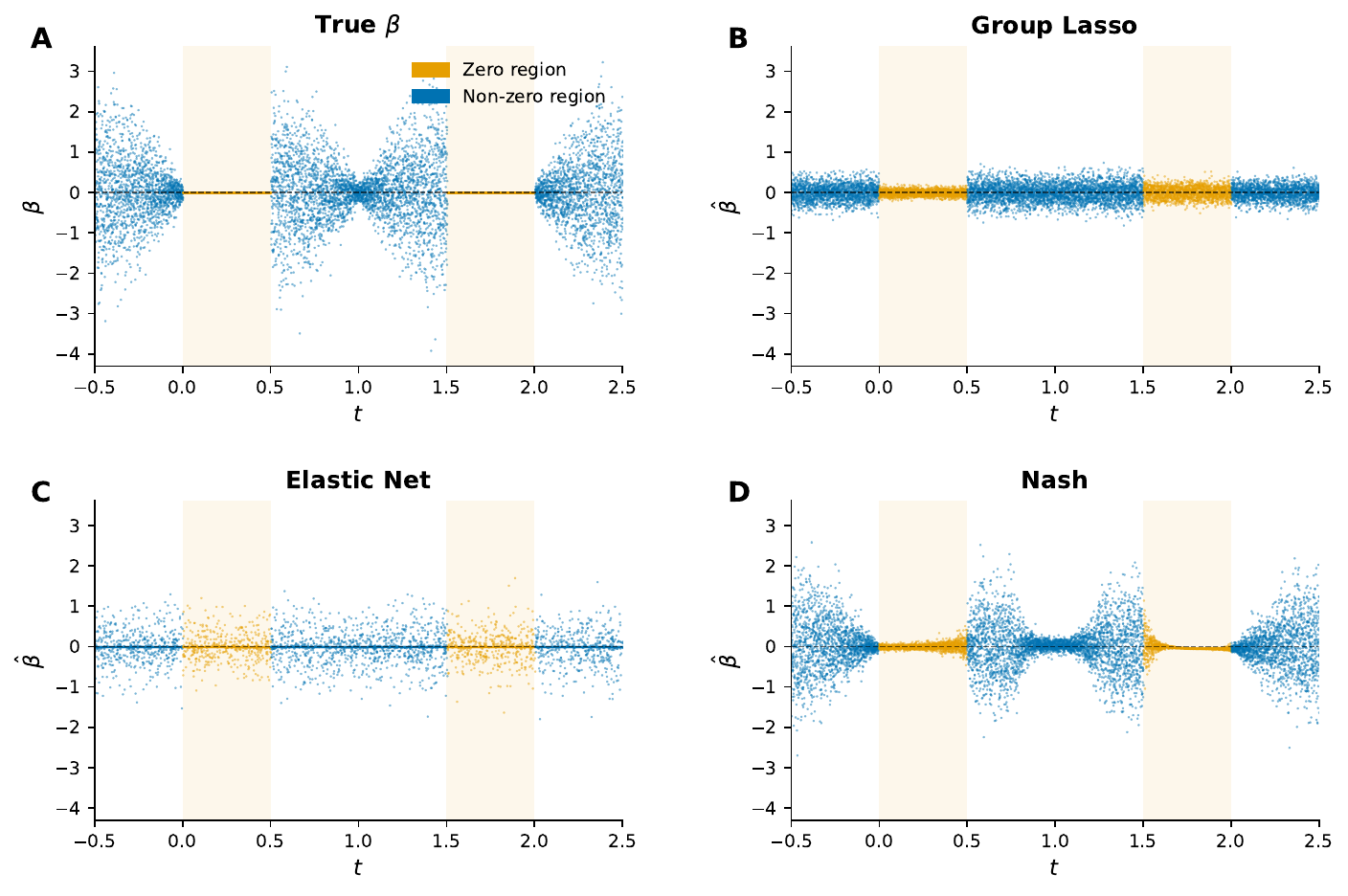}
  \caption{ Qualitative comparison of Group Lasso, Elastic Net, and Nash. 
Panel A shows the true $\beta$ in \textbf{continuous-index} simulations, with orange and blue points denoting structurally
zero and non-zero regions. Panels B--D show estimates from each method. Group Lasso, require a hand-crafted group indicator, whereas Nash (D) uses only the continuous index $t$ as side
information.}  \label{fig:illustration} 
\vspace{-2em}
\end{wrapfigure}
Following \citet{kim_flexible_2024}, we varied five factors: sparsity level $s$, signal-to-noise
ratio (PVE), number of predictors $p$, sample size $n$, and signal distribution $h$. We
considered two design types: independent Gaussian columns, equicorrelated Gaussian
columns with $\rho \in [0, 0.99]$. Each
setting is replicated 20 times (100 for the correlation sweep); full details are given in
Appendix~\ref{app:simdesign}. We summarize the results of these experiments in figure \ref{fig:placeholder} and supplementary figures \ref{fig:placeholder1}. Overall, Nash matches mr.ash, the current state-of-the-art, across nearly 
all simulation settings, outperforms it in several correlated-predictor  settings, and trails it only marginally in two to three settings (see Figure \ref{fig:placeholder}. Prediction accuracy is measured by the scaled RMSE as in \cite{kim_flexible_2024} (see equation \ref{eq:scale_rmse} in the appendix)\footnote{All confidence intervals reported in
this paper are 95\% Gaussian confidence intervals of the form
$\bar{x} \pm 1.96\,\widehat{\mathrm{SE}}(\bar{x})$, where $\bar{x}$ is
the mean of the per-replicate and $\widehat{\mathrm{SE}}$ is
its empirical standard error across replicates.}.

\subsection{Synthetic Experiments with Side Information}
\label{sec:synth_sideinfo}

\paragraph{Setup.} We design two controlled experiments. In the 
\textbf{known-groups} setting, $\mathbf{X} \in \mathbb{R}^{n \times p}$ 
with $n=500$ and $p \in \{100, 500, 1\,000, 10\,000\}$ predictors is 
partitioned into five groups with distinct sparsity and signal profiles 
(Table~\ref{tab:groups} , Appendix~\ref{app:synth_sideinfo}); side 
information $\mathbf{d}_j \in \{0,1\}^5$ is the one-hot group encoding. 
In the \textbf{continuous-index} setting, each predictor is assigned 
$t_j \sim \mathrm{Unif}(0,3)$ with spatially varying spike-and-slab 
coefficients and side information 
$\mathbf{d}_j = t_j $. 
In both settings, we  also compared mr.ash, Lasso, , elastic-net, group-Lasso and ipf-Lasso and for the \textit{continuous-index}, simulations group Lasso and
ipf-Lasso, require a hand-crafted group indicator, whereas
Nash uses only the continuous index $t$ as side information, we display the effect distribution along the covariate in Figure \ref{fig:illustration} panel A.  We also conduct an \textbf{ablation} comparing: 
 Nash-mdn   (using a full MLP prior of the form \ref{Nashmdn}), 
 Nash-linear  (single affine layer presented in paragraph \textbf{Group-Based and Hierarchical Penalty} section \ref{sec:plrandbeyond}, i.e., is the neural network 
necessary?),  Nash-no-cov  (no side information, is side 
information necessary?), and Nash-shuffled  (using Nash-mdn with shuffled side information, i.e., can noise covariate hinder the performance compared to no covariates?). We detail our experiments  in 
Appendix~\ref{app:synth_sideinfo}.

In both settings, \textsc{Nash} (full MLP prior) substantially outperforms  competing methods and its no-side-information counterpart (Tables~\ref{tab:ablation_groups} and~\ref{tab:ablation_c}). \textsc{Nash}-linear matches the full MLP in the \textit{known-groups} setting but degrades in the \textit{continuous-index} setting, where the affine model cannot recover the generative process, though it remains competitive with \textsc{Nash}-no-cov; \textsc{Nash}-shuffled likewise performs on par with \textsc{Nash}-no-cov, confirming robustness to noisy covariates. Notably, \textsc{Nash}-no-cov and mr.ash outperform ipf-Lasso and group-Lasso even in the known-group setting, which we attribute to the inability of L1 and L2 penalties methods to adapt to dense signal regimes—both mr.ash and \textsc{Nash} accommodate the full spectrum from sparse to dense effects.

 \subsection{Computational Cost of the M-step}\label{subsec:mstep}
\begin{wraptable}{r}{0.42\textwidth}
\vspace{-1em}
\centering
\caption{Mean M-step time (seconds) over 20 repeats.}
\label{tab:runtime}
\small
\begin{tabular}{rrrr}
\toprule
$p$ & CAVI  & Split VEB   & Speedup \\
\midrule
100     & 1.927   & 0.026 & $74\times$  \\
500     & 15.658  & 0.164 & $95\times$  \\
1\,000  & 34.582  & 0.363 & $95\times$  \\
10\,000 & 264.281 & 2.504 & $106\times$ \\
\bottomrule
\end{tabular}
\vspace{-3em}
\end{wraptable}

We compare the overall wall-clock time of the M-step for fitting the target model using the coordinate of \citep{kim_flexible_2024} and Nash  (split VEB) as a function of the number of predictors $p$. Since the E-step costs $\mathcal{O}(np)$ 
for both methods, we isolate the M-step wall-clock time, which is the computational bottleneck unique to each approach: $\mathcal{O}(p)$ sequential neural network backward passes for standard CAVI\citep{kim_flexible_2024} versus a 
single batched pass for the split-VEB of Nash.

\begin{wrapfigure}{r}{0.4\textwidth}
  \centering 
  \includegraphics[width=0.4\textwidth]{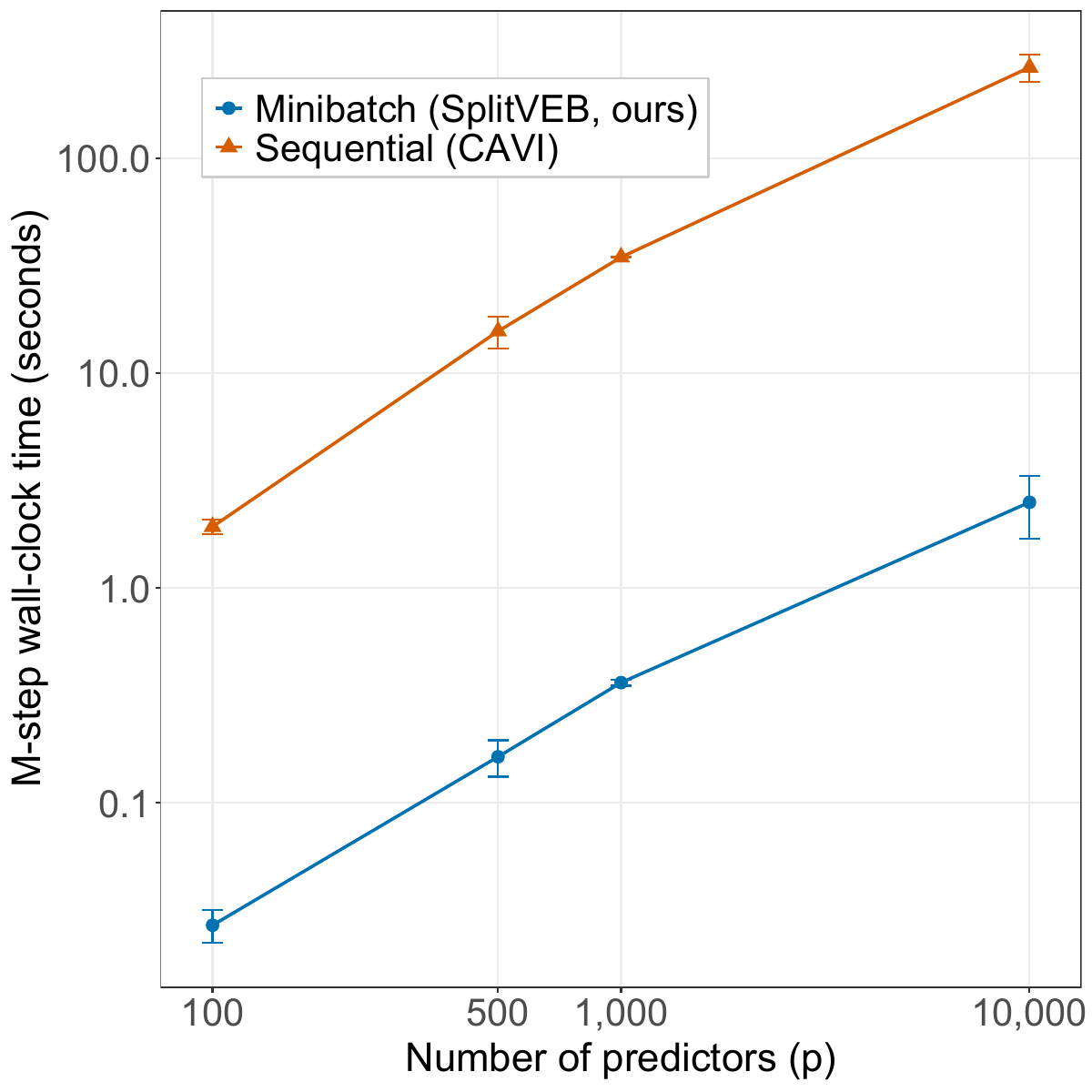}
  \caption{ M-step wall-clock time (log-log scale) versus number of predictors $p$ for direct CAVI \citep{kim_flexible_2024} and  Nash  (split VEB), averaged over 20 repeats on a single CPU core.  }  \label{fig:run_time} 
\vspace{-2em}
\end{wrapfigure}
Experiments are conducted on a single CPU core using the continuous-index simulation of Section~5.2 (one-dimensional side information), with $n = 500$ and $p \in \{100, 500, 1{,}000, 10{,}000\}$. The direct CAVI M-step processes effect estimates sequentially, performing $p$ forward backward passes one at a time; the Nash  M-step processes all $p$ estimates in a single batched forward-backward pass. Mean wall-clock times over 20 repeats are reported in Table~\ref{tab:runtime} and Figure~\ref{fig:run_time}.  Nash  achieves a $74\times$ 
speedup over direct CAVI at $p=100$, growing to $105\times$ at 
$p=10\,000$. The \textsc{Nash} M-step scales approximately linearly 
in $p$ (from $0.026$s to $2.504$s), consistent with the $\mathcal{O}(p)$ 
cost of a single batched pass over $p$ inputs. The coordinate ascend of \cite{kim_flexible_2024} seems to scale slightly faster than linear (from $1.927$s to $264.281$s), 
reflecting the $\mathcal{O}(p)$ sequential backward passes that 
cannot be parallelized. At $p=10\,000$, direct CAVI requires over 
four minutes for a single M-step, rendering it impractical for 
high-dimensional applications.

This comparison is \textbf{conservative} in two respects: the  Nash  M-step runs the cEBNM solver for multiple epochs (here 20), whereas direct CAVI performs only a single gradient step per coordinate; and the side information is kept as simple as possible (one-dimensional).  Nash  thus performs strictly more optimization work per sweep while remaining orders of magnitude faster.
  
\subsection{Real Data Experiments with Side Information}

We evaluate Nash on four datasets spanning small ($n \times p = 144 \times 144$) to 
large-scale ($679 \times 489{,}503$) regimes, with three types of side information: 
group membership, continuous time, and genomic annotations. Dataset details and 
preprocessing are given in Appendix~\ref{supsec:realdata} and described in Table \ref {table:datset_desc}. In all experiments, 20\% 
of samples are held out for testing and performance is measured by RMSE. Baselines 
include mr.ash, Lasso, Enet ($\alpha{=}0.5$), Ridge, XGBoost, MLP, ipf-Lasso (group 
datasets only), Nash without side information, and Nash-MDN (Eq.~\ref{Nashmdn}).

\begin{table*}[ht]
\centering
\scriptsize % or \small for slightly larger text
\resizebox{\textwidth}{!}{%
\begin{tabular}{lcccc }
\toprule
\textbf{Method} & \textbf{SNP500} & \textbf{Airpassenger} & \textbf{GSE40279} & \textbf{TCGA}   \\
\midrule
Dimension ($n\times p$) & $235 \times 85$ & $144 \times 144$ & $679 \times 489{,}503$ & $1{,}212 \times 18{,}300$   \\
Side info & group & time & probe type & pathway   \\
\midrule
Ridge        & $0.070$ $(0.065;0.076)$   & $33.0$ $(31.0;35.0)$  & $7.21$ $(6.83;7.58)$   & $0.536$ $(0.472;0.599)$   \\
Enet         & $0.071$ $(0.066;0.078)$   & $30.2$ $(29.5;30.9)$  & $5.28$ $(5.02;5.56)$   & $0.466$ $(0.411;0.522)$  \\
Lasso        & $0.093$ $(0.087;0.100)$   & $49.2$ $(48.7;49.7)$  & $5.39$ $(5.08;5.71)$   & $0.465$ $(0.412;0.518)$   \\
mr.ash       &  $0.082$ $(0.076;0.088)$  &$20.2$ $(19.1;21.3)$                 &  $ 5.25$ $(5.01;5.71)$                   & $0.449$ $(0.405;0.493)$                        \\
XGBoost      & $0.062$ $(0.053;0.069)$   & $18.2$ $(17.2;19.1)$  & $6.17$ $(5.83;6.52)$   & $0.549$ $(0.494;0.604)$  \\
MLP          & $0.441$ $(0.361;0.521)$   & $59.5$ $(58.3;60.8)$  & $13.62$ $(13.08;14.15)$ & $0.457$ $(0.349;0.566)$   \\
ipf-Lasso    & $0.066$ $(0.060;0.072)$   & NA                    & $ 5.06 $ $(4.81;5.33)$  & $0.443$ $(0.393;0.473)$      \\
Nash.no.cov  &$0.084$ $(0.079;0.089)$    &  $19.7$ $(18.6;20.7)$  &   $ 5.27$ $(5.01;5.53)$                       & $0.457$ $(0.412;0.504)$   \\
Nash.mdn     & $\mathbf {0.058}$ $(0.053;0.0643)$  &  $\mathbf {17.7}$ $(17.1;18.2)$                   & $ \mathbf {5.03}$ $( 4.68;5.38)$            &    $\mathbf{0.435}$ $(0.387;0.483)$         \\
\bottomrule
\end{tabular}%
}
\caption{Comparison of methods across datasets using RMSE. Parentheses denote 95\% confidence intervals based on Gaussian approximations. ipf-Lasso cannot handle time as side information, so we put NA for the Airpassenger experiment.}
\label{tab:method_comparison}
\vspace{-2em}
\end{table*}

\subsection{Denoising MNIST Images}

We evaluate the performance of Nash-fused \ref{eq:Nash-fused} using  a simple 2-layer message passing GNN to remove Gaussian noise in images compared to  methods for image denoising that only use a \textbf{single image} (as opposed to models trained on other images like diffusion model), such  as fused-Lasso   \citep{tibshirani_sparsity_2005}, Total Variation (TV) Denoising \citet{chambolle_algorithm_2004}, Non-Local Means (NLM ) \citep{buades_non-local_2005}, Gaussian Filtering \citep{gonzalez_digital_2002},
Median Filtering \citep{huang_fast_1979},  and Noise2Self \citep{batson_noise2self_2019}  of denoising noisy grayscale images from the MNIST dataset. Nash-fused was run treating each image as a 2D grid graph, where each pixel is a node connected to its 4-nearest neighbors, and Noise2Self was run using a convolutional neural net, which is substantially slower than Nash-fused. The true signal is the clean MNIST digit image scaled to [0,1], and additive Gaussian noise with  standard deviation of $\sigma =0.2$ was applied to produce the observed noisy image. The experiment is repeated over 100 randomly selected MNIST images. For each method, we report the root mean squared error (RMSE) between the denoised image and the ground truth. Results are summarized as boxplots in supplementary Figure \ref{fig:cnn}  and showcase an example in Figure \ref{fig:two_stack}. Additional examples of denoised images using Nash-fused can be found in the Appendix (see figure\ref{fig:mnist2}-\ref{fig:mnist8}).

 \begin{wrapfigure}{r}{0.5\textwidth}
  \centering
  \vspace{-1em}
  \includegraphics[width=0.48\textwidth]{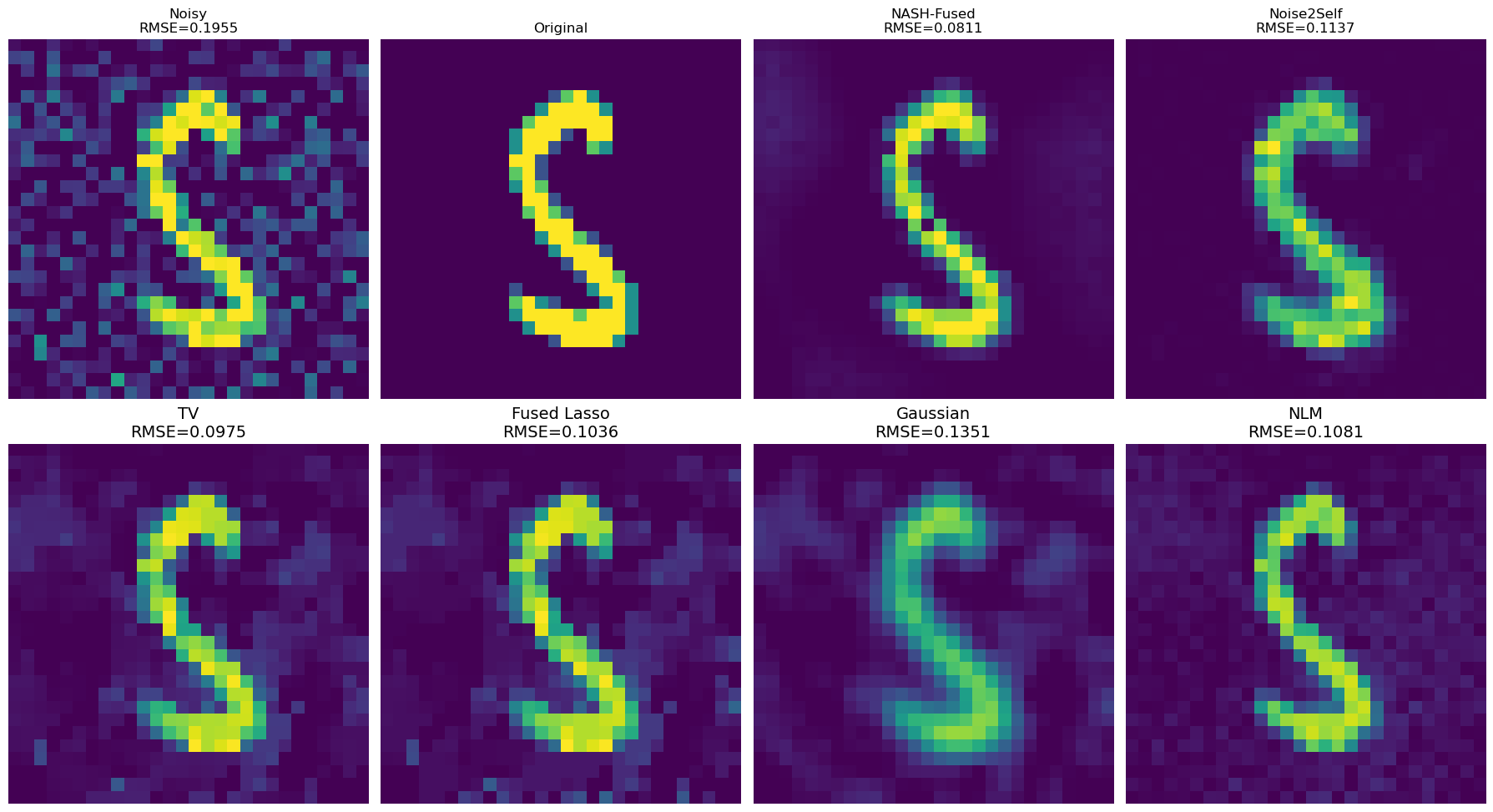}
  \caption{Sample from the experiment showcasing the performance of Nash-fused.}
  \label{fig:two_stack}
  \vspace{-1em}
\end{wrapfigure}

%% file: appendix.tex
%%%%%%%%%%%%%%%%%%%%%%%%%%%%%%%%%%%%%%%%%%%%%%%%%%%%%%%%%%%
\appendix
\onecolumn
\section{Appendix}
%%%%%%%%%%%%%%%%%%%%%%%%%%%%%%%%%%%%%%%%%%%%%%%%%%%%%%%%%%%

% ==================================================================
\subsection{Detailed Split VEB for the Nash Model}
\label{section:detailed}

The Nash model with side information is
\begin{align}
    \bm{y} \mid \bm{X}, \bm{\beta}, \sigma^2
        &\sim \mathcal{N}(\bm{X}\bm{\beta},\, \sigma^2 \bm{I}),
        \label{eq:nash_lik}\\
    \beta_j
        &\sim \mathcal{N}(b_j,\, \sigma_0^2),
        \label{eq:nash_layer}\\
    b_j
        &\sim g(\,\cdot\,;\, \bm{d}_j, \bm{\theta}).
        \label{eq:nash_prior}
\end{align}
We restrict the variational search to the mean-field family
$q(\bm{\beta}, \bm{b}) = \prod_{j=1}^{p} q_{\beta_j}(\beta_j)\, q_{b_j}(b_j)$.
The corresponding ELBO is
\begin{align}\label{eq:ELBO}
    \mathcal{F}(q_{\bm{\beta}},\, q_{\bm{b}},\, g;\,
               \sigma^2,\, \sigma_0^2)_{\mathrm{Nash}}
    =\;&
    \sum_i \mathbb{E}\bigl[\log p(y_i \mid \bm{x}_i, \bm{\beta}, \sigma^2)\bigr]
    +\\
    &\sum_j \mathbb{E}_{q_{\beta_j}q_{b_j}}\biggl[\log
        \frac{p(\beta_j \mid b_j, \sigma_0^2)}{q_{\beta_j}(\beta_j)}\biggr]
    + \sum_j \mathbb{E}\biggl[\log
        \frac{g(b_j;\, \bm{d}_j, \bm{\theta})}{q_{b_j}(b_j)}\biggr],
\end{align}
where the second term is taken jointly over $q_{\beta_j}$ and $q_{b_j}$
since $p(\beta_j \mid b_j, \sigma_0^2)$ depends on $b_j$.
The coordinate-ascent algorithm alternates between updating $q_{\bm{\beta}}$
and updating the pair $(q_{\bm{b}},\, g)$.

% ------------------------------------------------------------------
\subsubsection{Update for \texorpdfstring{$q_{\beta_j}$}{q\_beta\_j}}
\label{subsec:update_beta}

With $q_{\bm{b}}$ and $g$ fixed, the ELBO terms that depend on $q_{\beta_j}$ are
\begin{equation}
    \mathcal{F}(q_{\beta_j})_{\mathrm{Nash}}
    = \mathbb{E}\bigl[\log p(\bar{\bm{r}}_j \mid \bm{x}_j, \beta_j, \sigma^2)\bigr]
    + \mathbb{E}\bigl[\log p(\beta_j \mid \bar{b}_j, \sigma_0^2)\bigr]
    - \mathbb{E}\bigl[\log q_{\beta_j}(\beta_j)\bigr],
\end{equation}
where $\bar{\bm{r}}_j = \bm{y} - \bm{X}_{-j}\bar{\bm{\beta}}_{-j}$ is the
partial residual.  This reduction follows from the same  factorization
as Proposition~\ref{prop:cEBNM_reduction} (Section~\ref{sec:cEBNM_reduction}):
replacing the full $n$-dimensional likelihood by the scalar OLS statistic
$\hat{\beta}_{j,\text{MLE}} \colonequals \bm{x}_j^\top \bar{\bm{r}}_j/(n-1)$
introduces only a constant that is independent of $\beta_j$.  The optimal factor
$q^*_{\beta_j}$ is then the posterior of the conjugate sub-model
\begin{align}
    \bar{\bm{r}}_j &= \bm{x}_j \beta_j + \varepsilon,
    \quad \varepsilon \sim \mathcal{N}(\bm{0}, \sigma^2 \bm{I}),\\
    \beta_j &\sim \mathcal{N}(\bar{b}_j,\, \sigma_0^2),
\end{align}
namely $q^*_{\beta_j} = \mathcal{N}(\bar{\beta}_j,\, s_j^2)$ with
\begin{equation}
    s_j^2
    = \left(\frac{n-1}{\sigma^2}
            + \frac{1}{\sigma_0^2}\right)^{\!-1},
    \qquad
    \bar{\beta}_j
    = s_j^2\!\left(\frac{\bm{x}_j^\top \bar{\bm{r}}_j}{\sigma^2}
                + \frac{\bar{b}_j}{\sigma_0^2}\right)
    = \omega\,\hat{\beta}_{j,\text{MLE}} + (1-\omega)\,\bar{b}_j,
    \label{eq:beta_update}
\end{equation}
where $\omega \colonequals (n-1)\sigma_0^2/(\sigma^2 + (n-1)\sigma_0^2)
\in (0,1)$.  Thus $\bar{\beta}_j$ is a convex combination of the MLE
$\hat{\beta}_{j,\text{MLE}}$ and the current prior mean $\bar{b}_j$, weighted
by the relative precision of the likelihood versus the prior.
Each update requires $\mathcal{O}(n)$ operations.

% ------------------------------------------------------------------
\subsubsection{Update for \texorpdfstring{$q_{\bm{b}}$}{q\_b} and
\texorpdfstring{$g$}{g}}
\label{subsec:update_g}

Given $q_{\bm{\beta}}$, the terms in the ELBO depending on $(q_{\bm{b}}, g)$ are
\begin{equation}\label{eq:cebnm_elbo}
    \mathcal{F}(q_{\bm{b}},\, g)_{\mathrm{Nash}}
    = \sum_j \mathbb{E}_{q_{\beta_j}q_{b_j}}\bigl[\log p(\beta_j \mid b_j, \sigma_0^2)\bigr]
    + \sum_j \mathbb{E}\biggl[\log
        \frac{g(b_j;\, \bm{d}_j, \bm{\theta})}{q_{b_j}(b_j)}\biggr].
\end{equation}
Since $q_{\beta_j} = \mathcal{N}(\bar{\beta}_j, s_j^2)$ is held fixed in this
step, we integrate it out of the first term.  For any fixed $b_j$:
\begin{equation}
    \mathbb{E}_{q_{\beta_j}}\bigl[\log\mathcal{N}(\beta_j;\,b_j,\,\sigma_0^2)\bigr]
    = -\frac{\mathbb{E}_{q_{\beta_j}}[(\beta_j - b_j)^2]}{2\sigma_0^2} + \text{const}
    = -\frac{s_j^2 + (\bar{\beta}_j - b_j)^2}{2\sigma_0^2} + \text{const},
    \label{eq:qbeta_integrated}
\end{equation}
where we used $\mathbb{E}_{q_{\beta_j}}[(\beta_j - b_j)^2]
= \mathbb{E}_{q_{\beta_j}}[(\beta_j - \bar\beta_j)^2] + (\bar\beta_j - b_j)^2
= s_j^2 + (\bar\beta_j - b_j)^2$.
Taking the further expectation over $q_{b_j}$ and dropping $s_j^2$
(which is constant with respect to $q_{\bm{b}}$ and $g$):
\begin{equation}\label{eq:cebnm_reduced}
    \mathcal{F}(q_{\bm{b}},\, g)_{\mathrm{Nash}}
    \propto
    -\sum_j \frac{\mathbb{E}_{q_{b_j}}[(b_j - \bar{\beta}_j)^2]}{2\sigma_0^2}
    + \sum_j \mathbb{E}_{q_{b_j}}\!\biggl[\log\frac{g(b_j;\,\bm{d}_j,\bm{\theta})}
                                                     {q_{b_j}(b_j)}\biggr].
\end{equation}
This is precisely the ELBO of a \emph{covariate-moderated empirical Bayes
normal means} (cEBNM) problem with pseudo-observations $\bar{\beta}_j$ and
noise variance $\sigma_0^2$, confirming that $\bar{\beta}_j$ is the correct
sufficient statistic for this step
(see also \citealt{stephens_false_2017,willwerscheid_ebnm_2024}).

\paragraph{The cEBNM problem.}
\label{sec:ebnm}
Given pseudo-observations
$\bar{\beta}_j \overset{\mathrm{ind.}}{\sim} \mathcal{N}(b_j,\,\sigma_0^2)$
and the covariate-moderated prior
$b_j \overset{\mathrm{ind.}}{\sim} g(\bm{d}_j, \bm{\theta}) \in \mathcal{G}$,
the cEBNM problem requires two computations.
\begin{enumerate}
\item \emph{Parameter estimation:}
\begin{equation}\label{eq:cEBNM_mle}
    \hat{\bm{\theta}}
    = \operatorname*{arg\,max}_{\bm{\theta}}
      \prod_{j=1}^{p}
      \int \mathcal{N}(\bar{\beta}_j;\, b_j,\, \sigma_0^2)\;
           g(b_j;\, \bm{d}_j, \bm{\theta})\; db_j.
\end{equation}
\item \emph{Posterior summaries:}
\begin{equation}\label{eq:cEBNM_post}
    q_{b_j}^*(b_j)
    \;\propto\;
    \mathcal{N}(\bar{\beta}_j;\, b_j,\, \sigma_0^2)\;
    g(b_j;\, \bm{d}_j, \hat{\bm{\theta}}).
\end{equation}
\end{enumerate}
We write $\operatorname{cEBNM}(\bar{\bm{\beta}},\, \sigma_0^2,\, \bm{D})
= (\hat{\bm{\theta}},\, q_{\bm{b}})$.
Any prior family $\mathcal{G}$ for which both steps are computable is admissible. We refer to \cite{denault_covariate-moderated_2025} for more details on cEBNM solvers.

% ------------------------------------------------------------------
\subsubsection{Updates for \texorpdfstring{$\sigma^2$}{sigma2} and
\texorpdfstring{$\sigma_0^2$}{sigma02}}
\label{subsec:update_sigma}

Both variance parameters are updated by maximizing the ELBO while holding
$(\bar{\beta}_j, s_j^2, \bar{b}_j, V_{q_{b_j}})$ fixed.  The relevant ELBO terms are
\begin{align}
  \mathcal{F}(\sigma^2,\sigma_0^2)
  = -\frac{n}{2}\log\sigma^2
  - \frac{\mathbb{E}_q[\|\bm{y} - \bm{X}\bm{\beta}\|^2]}{2\sigma^2}
  - \frac{p}{2}\log\sigma_0^2
  - \frac{\sum_j \mathbb{E}_q[(\beta_j - b_j)^2]}{2\sigma_0^2},
\end{align}
where, using $\bm{x}_j^\top\bm{x}_j = n-1$ and mean-field independence:
\begin{align}
  \mathbb{E}_q[\|\bm{y} - \bm{X}\bm{\beta}\|^2]
  &= \|\bm{y} - \bm{X}\bar{\bm{\beta}}\|^2 + (n-1)\sum_j s_j^2,\\
  \mathbb{E}_q[(\beta_j - b_j)^2]
  &= (\bar{\beta}_j - \bar{b}_j)^2 + s_j^2 + V_{q_{b_j}}.
\end{align}
Setting the partial derivatives to zero:
\begin{align}
  (\sigma^2)^*
  &= \frac{1}{n}\Bigl(\|\bm{y} - \bm{X}\bar{\bm{\beta}}\|_2^2
     + (n-1)\sum_{j=1}^{p} s_j^2\Bigr),
  \label{eq:sigma2_update}\\[4pt]
  (\sigma_0^2)^*
  &= \frac{1}{p} \sum_{j=1}^{p}
      \Bigl(s_j^2 + (\bar{\beta}_j - \bar{b}_j)^2
            + V_{q_{b_j}}\Bigr).
  \label{eq:sigma02_update}
\end{align}

% ==================================================================
\subsection{Algorithm}
\label{sec:algorithm}

\begin{algorithm}[H]
\caption{Split-VEB coordinate ascent for Nash.}
\label{alg:Nash_CAVI}
\begin{algorithmic}
\Require
  Data $\bm{X}\!\in\!\mathbb{R}^{n\times p}$,
       $\bm{y}\!\in\!\mathbb{R}^n$;
  prior model $g(\,\cdot\,,\bm{\theta})$;
  initial estimates $\bar{\bm{\beta}},\bar{\bm{b}},\sigma^2,\sigma_0^2$.
\State $t \gets 0$;
  $\omega_0 \gets (n-1)\sigma_0^2/(\sigma^2+(n-1)\sigma_0^2)$
\Repeat
  \For{$j = 1$ \textbf{to} $p$}
    \State $\bar{\bm{r}}_j \gets \bar{\bm{r}} + \bm{x}_j \bar{\beta}_j^{\,t}$
        \Comment{Restore $j$-th effect}
    \State $\hat{\beta}_{j,\text{MLE}}^{\,t+1}
        \gets (\bm{x}_j^\top \bar{\bm{r}}_j)/(n-1)$
        \Comment{OLS sufficient statistic}
    \State $\bar{\beta}_j^{\,t+1}
        \gets \omega_t\,\hat{\beta}_{j,\text{MLE}}^{\,t+1}
              + (1-\omega_t)\,\bar{b}_j^{\,t}$
        \Comment{Posterior mean }
    \State $\bar{\bm{r}} \gets \bar{\bm{r}}_j - \bm{x}_j\,\bar{\beta}_j^{\,t+1}$
  \EndFor
  \State $\hat{\bm{\theta}}, q_{\bm{b}}^{t+1}
      \gets \operatorname{cEBNM}\!\left(
        \bar{\bm{\beta}}^{\,t+1},\, \sigma_{0,t}^2,\, \bm{D}\right)$
        \Comment{Global M-step: single neural-network update}
  \State $\bar{\bm{b}}^{\,t+1}
      \gets \mathbb{E}_{q_{\bm{b}}^{t+1}}[\bm{b}]$
  \State $(\sigma^2)^{t+1}
      \gets$ closed form~\eqref{eq:sigma2_update}
  \State $(\sigma_0^2)^{t+1}
      \gets$ closed form~\eqref{eq:sigma02_update}
  \State $\omega_{t+1}
      \gets (n-1)(\sigma_0^2)^{t+1}\,/\,
             \bigl(\sigma_{t+1}^2 + (n-1)(\sigma_0^2)^{t+1}\bigr)$
  \State $t \gets t + 1$
\Until{convergence}
\State \Return $\bar{\bm{b}},\; \hat{\bm{\theta}},\; \sigma^2$
\end{algorithmic}
\end{algorithm}

\paragraph{Computational cost.}
Each inner-loop iteration costs $\mathcal{O}(n)$ (a dot product).  The full
E-step over all $j$ costs $\mathcal{O}(np)$.  The M-step (cEBNM solve) requires
a \emph{single} forward--backward pass through the neural network, regardless
of $p$.  This reduces the neural-network cost per sweep from $\mathcal{O}(p)$
passes (as would be required by direct CAVI on the target model; see
Section~\ref{sec:relaxation}) to $\mathcal{O}(1)$.

% ==================================================================
\subsection{Nash as a Continuous Relaxation of the Target Model}
\label{sec:relaxation}

We clarify the relationship between Nash and the target model
(Section~\ref{subsec:target_model}), and identify the computational bottleneck
that splits VEB is designed to circumvent.

\paragraph{Direct CAVI on the target model.}
By Corollary~\ref{cor:target_cEBNM} (Section~\ref{sec:cEBNM_reduction}), direct
coordinate ascent on the target model ELBO reduces to a cEBNM driven by the OLS
sufficient statistics $\hat{\beta}_{j,\text{MLE}}$:
\begin{equation}\label{eq:target_cebnm}
  \hat{\bm{\theta}}_{\text{target}}
  = \operatorname*{arg\,max}_{\bm{\theta}}
    \prod_{j=1}^p
    \int \mathcal{N}\!\left(\hat{\beta}_{j,\text{MLE}};\,\beta_j,\,s_j^2\right)
    g(\beta_j;\,\bm{d}_j,\,\bm{\theta})\,d\beta_j,
\end{equation}
with $s_j^2 = \sigma^2/(n-1)$.  This is exact: it optimizes the target model
ELBO directly, and the neural network appears only in the M-step.  However, the
M-step must be re-run after \emph{every coordinate update} because each
$\hat{\beta}_{j,\text{MLE}}$ changes when any other $\bar{\beta}_{j'}$ changes,
requiring $p$ neural-network backward passes per sweep --- prohibitive for large $p$.

\paragraph{How Nash circumvents this bottleneck.}
Nash introduces the auxiliary layer $\beta_j \mid b_j \sim \mathcal{N}(b_j,
\sigma_0^2)$.  For $\sigma_0^2 > 0$, the M-step input is the shrinkage estimate
$\bar{\beta}_j = \omega\,\hat{\beta}_{j,\text{MLE}} + (1-\omega)\bar{b}_j$
rather than the raw OLS estimate.  Because $\bar{\beta}_j$ is a smooth function
of $\bar{b}_j$, all $p$ coordinates can be updated before the neural network is
retrained.  This decoupling permits a single global M-step per sweep.

The parameter $\omega = (n-1)\sigma_0^2/(\sigma^2+(n-1)\sigma_0^2)$ governs
the trade-off: large $\sigma_0^2$ means $\omega \approx 1$ and $\bar{\beta}_j
\approx \hat{\beta}_{j,\text{MLE}}$ (Nash behaves like direct CAVI but with a
decoupled M-step); small $\sigma_0^2$ means $\omega \approx 0$ and $\bar{\beta}_j
\approx \bar{b}_j$ (stronger coupling to the prior, tighter self-consistency at
the fixed poin).  Since $\sigma_0^2$ is estimated by maximizing the Nash ELBO,
the algorithm selects this trade-off automatically from the data.

\subsection{Nash Optimizes a Lower Bound of the Marginal CAVI ELBO}
\label{sec:connection_mrash}

We formalize the relationship between the Nash split-VEB algorithm and
standard coordinate-ascent variational inference applied to the Nash marginal
model.  Throughout, we work without side information, setting
$g(\bm{d}_j,\bm{\theta}) = g(\bm{\theta})$ for all $j$, and fix
$\bm{\beta}_{-j}$; the analysis extends to the full sweep coordinate-wise.

\paragraph{Setup.}
The \emph{Nash marginal likelihood} of $b_j$ is
\begin{equation}\label{eq:f_def}
    f(b_j)
    := p(\bm{r}_j \mid \bm{x}_j, b_j, \sigma^2, \sigma_0^2)
    = \int p(\bm{r}_j \mid \bm{x}_j, \beta_j, \sigma^2)\,
            \mathcal{N}(\beta_j;\, b_j,\, \sigma_0^2)\, d\beta_j,
\end{equation}
where $\bm{r}_j = \bm{y} - \bm{X}_{-j}\bm{\beta}_{-j}$.
Applying standard mean-field CAVI to the Nash model with $b_j$ as the
inference variable yields the \emph{Nash marginal CAVI ELBO}
\begin{equation}\label{eq:G_def}
    \mathcal{G}(q_{b_j}, g)
    := \mathbb{E}_{q_{b_j}}\!\bigl[\log f(b_j)\bigr]
         + \mathbb{E}_{q_{b_j}}\!\biggl[
             \log\frac{g(b_j)}{q_{b_j}(b_j)}\biggr].
\end{equation}
Since $f(b_j)$ is the Nash augmented model marginal likelihood (not the target
model likelihood), $\mathcal{G}$ is the ELBO of the Nash augmented model and
does not in general equal the target model ELBO.  %As $\sigma_0^2 \to 0$, however, $f(b_j) \to p(\bm{r}_j \mid \bm{x}_j, b_j, \sigma^2)$ and $\mathcal{G}$ converges to the target model ELBO (Proposition~\ref{prop:limit}).

Split VEB replaces the intractable expectation $\mathbb{E}[\log f(b_j)]$ by the
cheaper evaluation at the mean $\bar{b}_j$ (Lemma~\ref{lem:profiled}),
incurring a further approximation error that is bounded by the curvature of
$\log f$ (Theorem~\ref{thm:lower_bound}).

\begin{lemma}[Profiled Nash ELBO]\label{lem:profiled}
Let $\bar{b}_j = \mathbb{E}_{q_{b_j}}[b_j]$ and
$V_j = \operatorname{Var}_{q_{b_j}}(b_j)$.  For any $q_{b_j}$,
\begin{equation}\label{eq:profiled_nash}
    F(b_j, g;\, \sigma^2, \sigma_0^2)
    := \max_{q_{\beta_j}}
           \mathcal{F}_{\mathrm{Nash}}(q_{\beta_j}, q_{b_j}, g)
    = \log f(\bar{b}_j)
          + \mathbb{E}_{q_{b_j}}\!\biggl[
              \log\frac{g(b_j)}{q_{b_j}(b_j)}\biggr]
          - \frac{V_j}{2\sigma_0^2}.
\end{equation}
\end{lemma}

\begin{proof}
With $q_{b_j}$ held fixed (so $\bar{b}_j$ and $V_j$ are constants),
we first integrate $q_{b_j}$ out of the coupling term.  For any fixed $\beta_j$:
\begin{equation}
  \mathbb{E}_{q_{b_j}}\bigl[\log\mathcal{N}(\beta_j;\,b_j,\,\sigma_0^2)\bigr]
  = -\frac{(\beta_j - \bar{b}_j)^2 + V_j}{2\sigma_0^2} + \text{const}
  = \log\mathcal{N}(\beta_j;\,\bar{b}_j,\,\sigma_0^2) - \frac{V_j}{2\sigma_0^2}
    + \text{const},
  \label{eq:qb_integrated}
\end{equation}
where the identity uses
$\mathbb{E}_{q_{b_j}}[(b_j - \beta_j)^2] = (\bar{b}_j - \beta_j)^2 + V_j$.
The $-V_j/(2\sigma_0^2)$ term does not depend on $q_{\beta_j}$ and
factors out of the maximisation.  Substituting~\eqref{eq:qb_integrated}
into the Nash ELBO:
\begin{align}
  \mathcal{F}_{\mathrm{Nash}}(q_{\beta_j}, q_{b_j}, g)
  = &\mathbb{E}_{q_{\beta_j}}\!\left[
      \log p(\bm{r}_j \mid \bm{x}_j, \beta_j, \sigma^2)
      + \log\frac{\mathcal{N}(\beta_j;\,\bar{b}_j,\,\sigma_0^2)}
                 {q_{\beta_j}(\beta_j)}
    \right]
  + \\
  & \mathbb{E}_{q_{b_j}}\!\left[
      \log\frac{g(b_j)}{q_{b_j}(b_j)}
    \right]
  - \frac{V_j}{2\sigma_0^2}
  + \text{const}.
\end{align}
The first bracket is the ELBO for the model
$p(\bm{r}_j \mid \bm{x}_j, \beta_j, \sigma^2)\,\mathcal{N}(\beta_j;\,\bar{b}_j,\,\sigma_0^2)$;
it is maximised by
$q^*_{\beta_j} = p(\beta_j \mid \bm{r}_j,\,\bar{b}_j,\,\sigma^2,\,\sigma_0^2)$,
at which its value equals $\log f(\bar{b}_j)$ by the ELBO
identity.%~\eqref{eq:var_rep}.
The remaining terms do not depend on
$q_{\beta_j}$, giving~\eqref{eq:profiled_nash}.
\end{proof}

\begin{lemma}[Curvature lower bound]\label{lem:curvature}
$\displaystyle\frac{d^2 \log f(b_j)}{db_j^2} \geq -\frac{1}{\sigma_0^2}$
for all $b_j \in \mathbb{R}$ and $\sigma_0^2 > 0$.
\end{lemma}

\begin{proof}
Write $\mathbb{E}[\,\cdot\,]$ for expectations under the posterior
$p(\beta_j \mid \bm{r}_j, b_j, \sigma^2, \sigma_0^2)$.
Differentiating $f(b_j) = \int p(\bm{r}_j \mid \beta_j)
\mathcal{N}(\beta_j; b_j, \sigma_0^2)\, d\beta_j$ under the integral sign
and using the exponential-family score identity for
$\mathcal{N}(\beta_j; b_j, \sigma_0^2)$ as a natural exponential family in $b_j$:
\begin{equation}
\frac{d \log f(b_j)}{db_j}
    = \frac{\mathbb{E}[\beta_j] - b_j}{\sigma_0^2}.
    \label{eq:score}
\end{equation}
Differentiating once more and applying the standard identity
$d\,\mathbb{E}[\beta_j]/db_j = \operatorname{Var}(\beta_j)/\sigma_0^2$
(which follows from differentiating the posterior mean with respect to the
natural parameter $b_j/\sigma_0^2$):
\begin{equation}
\frac{d^2 \log f(b_j)}{db_j^2}
    = \frac{\operatorname{Var}(\beta_j)}{\sigma_0^4}
      - \frac{1}{\sigma_0^2}
    \;\geq\; -\frac{1}{\sigma_0^2},
\end{equation}
since $\operatorname{Var}(\beta_j) \geq 0$.
\end{proof}

\begin{theorem}\label{thm:lower_bound}
For any $q_{b_j}$, any prior $g$, and any $\sigma_0^2 > 0$:
$\;\mathcal{G}(q_{b_j}, g) \;\geq\; F(b_j, g;\, \sigma^2, \sigma_0^2)$.
\end{theorem}

\begin{proof}
By Lemma~\ref{lem:profiled} and definition~\eqref{eq:G_def}, the inequality
reduces to $\mathbb{E}_{q_{b_j}}[\log f(b_j)] \geq \log f(\bar{b}_j)
- V_j/(2\sigma_0^2)$.  By Taylor's theorem, for any $b_j$ in the support
of $q_{b_j}$,
\begin{equation}
    \log f(b_j)
    = \log f(\bar{b}_j)
    + \frac{d\log f}{db_j}\bigg|_{\bar{b}_j}\!(b_j - \bar{b}_j)
    + \frac{1}{2}
      \frac{d^2\log f}{db_j^2}\bigg|_{\xi(b_j)}
      (b_j - \bar{b}_j)^2,
\end{equation}
where $\xi(b_j)$ lies between $b_j$ and $\bar{b}_j$.  Taking expectations
under $q_{b_j}$, the linear term vanishes since
$\mathbb{E}_{q_{b_j}}[b_j - \bar{b}_j] = 0$, and
Lemma~\ref{lem:curvature} gives
$\mathbb{E}_{q_{b_j}}[\log f(b_j)] \geq \log f(\bar{b}_j) - V_j/(2\sigma_0^2)$.
\end{proof}

\begin{remark}[Approximation hierarchy]
\label{rem:hierarchy}
The Nash split-VEB algorithm optimizes the profiled ELBO
$F(b_j, g;\,\sigma^2,\sigma_0^2)$, which satisfies
\begin{equation}
  \underbrace{\mathcal{G}(q_{b_j}, g)}_{\substack{\text{Nash augmented}\\\text{model ELBO}}}
  \;\geq\;
  \underbrace{F(b_j, g;\,\sigma^2,\sigma_0^2)}_{\substack{\text{Nash split-VEB}\\\text{(profiled) ELBO}}},
\end{equation}
by Theorem~\ref{thm:lower_bound}.  Split VEB is therefore conservative
relative to the Nash marginal CAVI ELBO: it never overclaims evidence.
Both $\mathcal{G}$ and $F$ are ELBOs for the Nash augmented model and
approximate the target model ELBO. % with both approximations becoming exact as $\sigma_0^2 \to 0$ (Proposition~\ref{prop:limit}). 
The learned $\sigma_0^2$ controls the quality of both approximations simultaneously:
small $\sigma_0^2$ tightens both gaps toward the target; large $\sigma_0^2$
improves computational efficiency at the cost of approximation quality.
\end{remark}

% ==================================================================
\subsection{Reduction to Univariate cEBNM }
\label{sec:cEBNM_reduction}

We show that the coordinate-ascent updates for both the target model and the
Nash model reduces to cEBNM problems driven by scalar OLS statistics.

\begin{lemma} 
\label{lem:pyth_target}
Let $\hat{\beta}_{j,\text{MLE}} \colonequals \bm{x}_j^{\top}\bar{\bm{r}}_j/(n-1)$
and $s_j^2 \colonequals \sigma^2/(n-1)$.  Then
\begin{equation}
  \label{eq:pyth_target}
  \log p\!\left(\bar{\bm{r}}_j \mid \beta_j,\,\sigma^2\right)
  = \log \mathcal{N}\!\left(\hat{\beta}_{j,\text{MLE}};\;\beta_j,\;s_j^2\right)
  + C\!\left(\bar{\bm{r}}_j,\,\sigma^2\right),
\end{equation}
where $C$ does not depend on $\beta_j$.
\end{lemma}

\begin{proof}
Decompose $\bar{\bm{r}}_j - \bm{x}_j\beta_j
  = (\bar{\bm{r}}_j - \bm{x}_j\hat{\beta}_{j,\text{MLE}})
    + \bm{x}_j(\hat{\beta}_{j,\text{MLE}} - \beta_j)$.
The cross term $\bm{x}_j^{\top}(\bar{\bm{r}}_j - \bm{x}_j\hat{\beta}_{j,\text{MLE}})
= \bm{x}_j^{\top}\bar{\bm{r}}_j - (n-1)\hat{\beta}_{j,\text{MLE}} = 0$
by definition.  Hence, using $\bm{x}_j^{\top}\bm{x}_j = n-1$:
\begin{equation}
  \bigl\|\bar{\bm{r}}_j - \bm{x}_j\beta_j\bigr\|^2
  = \bigl\|\bar{\bm{r}}_j - \bm{x}_j\hat{\beta}_{j,\text{MLE}}\bigr\|^2
  + (n-1)(\hat{\beta}_{j,\text{MLE}} - \beta_j)^2.
\end{equation}
Dividing by $2\sigma^2$ gives~\eqref{eq:pyth_target}.
\end{proof}

\begin{proposition} 
\label{prop:cEBNM_reduction}
In the Nash model, integrating out $\beta_j$ gives
\begin{equation}
  \label{eq:marginal_bj}
  \log p\!\left(\bar{\bm{r}}_j \mid b_j,\,\sigma^2,\,\sigma_0^2\right)
  = \log \mathcal{N}\!\left(\hat{\beta}_{j,\text{MLE}};\; b_j,\; s_j^2 + \sigma_0^2\right)
  + C\!\left(\bar{\bm{r}}_j,\,\sigma^2,\,\sigma_0^2\right),
\end{equation}
where $C$ does not depend on $b_j$.  As $\sigma_0^2\to 0$,
\eqref{eq:marginal_bj} reduces exactly to \eqref{eq:pyth_target}.
\end{proposition}

\begin{proof}
 
Since $\beta_j \mid b_j \sim \mathcal{N}(b_j,\sigma_0^2)$ and
$\bar{\bm{r}}_j \mid \beta_j \sim \mathcal{N}(\bm{x}_j\beta_j, \sigma^2\bm{I}_n)$:
\begin{equation}
  \bar{\bm{r}}_j \mid b_j
  \sim \mathcal{N}\!\left(\bm{x}_j b_j,\;
    \boldsymbol{\Sigma}_j\right),
  \quad
  \boldsymbol{\Sigma}_j
  \colonequals
  \sigma^2\bm{I}_n + \sigma_0^2\bm{x}_j\bm{x}_j^{\top}.
\end{equation}

By Sherman--Morrison inversion we have that,  $\bm{x}_j^{\top}\bm{x}_j = n-1$:
\begin{equation}
  \boldsymbol{\Sigma}_j^{-1}
  = \frac{1}{\sigma^2}\bm{I}_n
  - \frac{\sigma_0^2}{\sigma^2\bigl(\sigma^2 + (n-1)\sigma_0^2\bigr)}
  \bm{x}_j\bm{x}_j^{\top}.
\end{equation}
  By the same token as in   Lemma~\ref{lem:pyth_target}:
\begin{equation}
  \bigl\|\bar{\bm{r}}_j - \bm{x}_j b_j\bigr\|^2
  = \bigl\|\bar{\bm{r}}_j - \bm{x}_j\hat{\beta}_{j,\text{MLE}}\bigr\|^2
  + (n-1)(\hat{\beta}_{j,\text{MLE}} - b_j)^2.
\end{equation}
 
Substituting the Sherman--Morrison inverse and the Pythagorean decomposition,
the coefficient of $(\hat{\beta}_{j,\text{MLE}} - b_j)^2$ in
$(\bar{\bm{r}}_j - \bm{x}_j b_j)^\top \boldsymbol{\Sigma}_j^{-1}
 (\bar{\bm{r}}_j - \bm{x}_j b_j)$ is
\begin{equation}
  \frac{n-1}{\sigma^2}
  - \frac{\sigma_0^2(n-1)^2}{\sigma^2(\sigma^2+(n-1)\sigma_0^2)}
  = \frac{(n-1)\sigma^2}{\sigma^2(\sigma^2+(n-1)\sigma_0^2)}
  = \frac{1}{s_j^2+\sigma_0^2}.
\end{equation}
The remaining term $\|\bar{\bm{r}}_j - \bm{x}_j\hat{\beta}_{j,\text{MLE}}\|^2/\sigma^2$
and the log-determinant $\log\det(2\pi\boldsymbol{\Sigma}_j)$ are both
independent of $b_j$.  This establishes~\eqref{eq:marginal_bj}.
Setting $\sigma_0^2=0$ recovers~\eqref{eq:pyth_target} with no
$1/\sigma_0^2$ blowup, since the $\sigma_0^2$ factors cancel.
\end{proof}

\begin{corollary}[Target model reduces to cEBNM]
\label{cor:target_cEBNM}
Under the target model, the profiled coordinate-ascent VEB objective for $g$
reduces to the cEBNM problem:
\begin{align}
  \hat{\bm{\theta}}
  &= \argmax_{\bm{\theta}}
  \prod_{j=1}^{p}
  \int
  \mathcal{N}\!\left(\hat{\beta}_{j,\text{MLE}};\;\beta_j,\;s_j^2\right)
  g(\beta_j;\,\bm{d}_j,\,\bm{\theta})\,d\beta_j,
  \label{eq:cEBNM_theta}\\
  q_{\beta_j}^{*}(\beta_j)
  &\propto
  \mathcal{N}\!\left(\hat{\beta}_{j,\text{MLE}};\;\beta_j,\;s_j^2\right)
  g(\beta_j;\,\bm{d}_j,\,\hat{\bm{\theta}}).
  \label{eq:cEBNM_post_target}
\end{align}
This requires updating~\eqref{eq:cEBNM_theta} after every change to any
$\hat{\beta}_{j',\text{MLE}}$, necessitating $p$ neural-network backward
passes per sweep.  Nash with $\sigma_0^2 > 0$ replaces the exact
$\hat{\beta}_{j,\text{MLE}}$ by the shrinkage estimate $\bar{\beta}_j$,
decoupling the M-step and reducing the cost to a single backward pass.
\end{corollary}

\begin{proof}
By Lemma~\ref{lem:pyth_target}, the ELBO at coordinate $j$ is
$\mathcal{F}(q_{\beta_j},\bm{\theta})
= \mathbb{E}[\log\mathcal{N}(\hat{\beta}_{j,\text{MLE}};\beta_j,s_j^2)]
- D_{\mathrm{KL}}(q_{\beta_j}\,\|\,g(\cdot;\bm{d}_j,\bm{\theta})) + C$.
maximizing over $q_{\beta_j}$ yields~\eqref{eq:cEBNM_post_target}; substituting
back and summing over $j$ gives~\eqref{eq:cEBNM_theta}.
\end{proof}

\begin{remark}
Proposition~\ref{prop:cEBNM_reduction} shows that Nash implements the same
cEBNM structure as Corollary~\ref{cor:target_cEBNM}, but with inflated noise
variance $s_j^2 + \sigma_0^2$ and the shrinkage estimate $\bar{\beta}_j$
replacing the exact MLE $\hat{\beta}_{j,\text{MLE}}$.  Both effects are
controlled by $\sigma_0^2$ and vanish as $\sigma_0^2 \to 0$, recovering exact
target-model inference at the cost of the decoupling benefit.
\end{remark}

%% file: experiments.tex
\section{Simulation Design}
\label{app:simdesign}

 \subsection{Comparison with baselines}

All data sets are generated using the simulation wrapper of \citet{kim_flexible_2024}, available
at \url{https://github.com/stephenslab/mr-ash-workflow}. For each replicate $i$, a
design matrix $\mathbf{X} \in \mathbb{R}^{n \times p}$ is drawn under one of three
designs described below, and the response is
\begin{equation}
  \mathbf{y} = \mathbf{X}\boldsymbol{\beta} + \boldsymbol{\varepsilon},
  \qquad
  \boldsymbol{\varepsilon} \sim \mathcal{N}(\mathbf{0}, \sigma^2 \mathbf{I}_n),
\end{equation}
where $\sigma^2$ is set so that
$\mathrm{Var}(\mathbf{X}\boldsymbol{\beta}) /
 [\mathrm{Var}(\mathbf{X}\boldsymbol{\beta}) + \sigma^2] = \mathrm{PVE}$.
Unless stated otherwise, $n = 500$, $s = 20$ non-zero coefficients are chosen
uniformly at random from $\{1,\ldots,p\}$, $\mathrm{PVE} = 0.5$, and the non-zero
signals are drawn i.i.d.\ from the standard normal distribution. Each held-out test
set is the same size as the training set and is generated from the same
$\boldsymbol{\beta}$. We summarize the simulation scenarios in Table \ref{tab:scenarios}

\begin{table}[ht]
\centering

\small
\begin{tabular}{ llll}
\toprule
  Varied factor & Fixed parameters & Design & Levels \\
\midrule
  Sparsity $s$        & $p=200$,\ PVE$=0.5$,\ $h=$Normal          & Indep       & $s\in\{1,2,5,10,20,50,100,200\}$ \\
  Sparsity $s$        & $p=10{,}000$,\ PVE$=0.5$,\ $h=$Normal     & Indep       & $s\in\{1,5,20,100,500,2000,10000\}$ \\
  Sparsity $s$        & $p=200$,\ PVE$=0.5$,\ $h=$Constant         & Indep       & $s\in\{1,2,5,10,20,50,100,200\}$ \\
  Sparsity $s$        & $p=200$,\ PVE$=0.9$,\ $h=$Normal           & Indep       & $s\in\{1,2,5,10,20,50,100,200\}$ \\
  Sparsity $s$        & $p=2{,}000$,\ PVE$=0.5$,\ $\rho=0.95$     & Equicorr    & $s\in\{1,5,20,100,500,2000\}$ \\

\midrule
  Sample size $n$     & $p=2{,}000$,\ $s=20$,\ PVE$=0.5$           & Indep       & $n\in\{200,500,1000,2000,5000\}$ \\
\midrule
  Signal strength     & $p=2{,}000$,\ $s=20$                        & Indep       & PVE$\in\{0.0,0.1,\ldots,0.9\}$ \\
\midrule
 Signal dist.\ $h$   & $p=200$,\ $s=200$,\ PVE$=0.5$ (dense)      & Indep       & $t(1,2,4,8)$; Lap.; Norm.; Unif.; Const. \\
  Signal dist.\ $h$   & $p=2{,}000$,\ $s=20$,\ PVE$=0.5$ (sparse)  & Indep       & same as above \\
\midrule
  Correlation $\rho$  & $p=2{,}000$,\ $s=20$,\ PVE$=0.5$           & Equicorr    & $\rho\in\{0,0.1,\ldots,0.9,0.95,0.99\}$ \\
  Correlation $\rho$  & $p=200$,\ $s=20$,\ PVE$=0.5$               & Equicorr    & same as above \\
\bottomrule
\end{tabular}
\caption{Summary of the  simulation scenarios. ``Equicorr'' denotes an equicorrelated
         Gaussian design with the stated $\rho$; ``Indep'' denotes independent
         standard Gaussian columns. Signal distributions are: Normal
         $\mathcal{N}(0,1)$, Constant (point mass), Laplace, Uniform on $[-1,1]$,
         and Student-$t$ with varying degrees of freedom.}
\label{tab:scenarios}
\end{table}

\noindent
All scenarios use $n=500$ unless stated otherwise. Scenarios~1--6 together reproduce
Experiment~1 of \citet{kim_flexible_2024} (varying sparsity) across different dimensionalities
and designs. Scenario~7 reproduces their Experiment~4 (varying $p$, here recast as
varying $n$ at fixed $p$). Scenario~8 reproduces their Experiment~2 (varying PVE).
Scenarios~9--10 reproduce their Experiment~3 (varying signal distribution) in a dense
and a sparse setting, respectively. Scenarios~11--12 complement the
correlated-predictor panels of their Experiment~1 with a finer grid over $\rho$, and
are replicated 100 times each to reduce Monte Carlo noise at high correlations.

 Prediction accuracy is measured by the scaled RMSE as in \cite{kim_flexible_2024}
\begin{equation}\label{eq:scale_rmse}
  \mathrm{RMSE}\text{-}\mathrm{scaled}
  \bigl(\mathbf{y}_{\mathrm{test}},\hat{\boldsymbol{\beta}}\bigr)
  \;\triangleq\;
  \frac{\mathrm{RMSE}\bigl(\mathbf{y}_{\mathrm{test}},\hat{\boldsymbol{\beta}}\bigr)}%
       {\mathrm{RMSE}(\hat{\boldsymbol{\beta}}=\mathbf{0})},
  \qquad
  \mathrm{RMSE}\bigl(\mathbf{y}_{\mathrm{test}},\hat{\boldsymbol{\beta}}\bigr)
  \;\triangleq\;
  \frac{1}{\sqrt{n}}
  \bigl\|\mathbf{y}_{\mathrm{test}}
         - \mathbf{X}_{\mathrm{test}}\hat{\boldsymbol{\beta}}\bigr\|_2
\end{equation}
and $\mathrm{RMSE}(\hat{\boldsymbol{\beta}}=\mathbf{0}) \triangleq \sigma/\sqrt{1-\mathrm{PVE}}$
is the expected RMSE of the null predictor.

\begin{figure}
    \centering
    \includegraphics[width=1\linewidth]{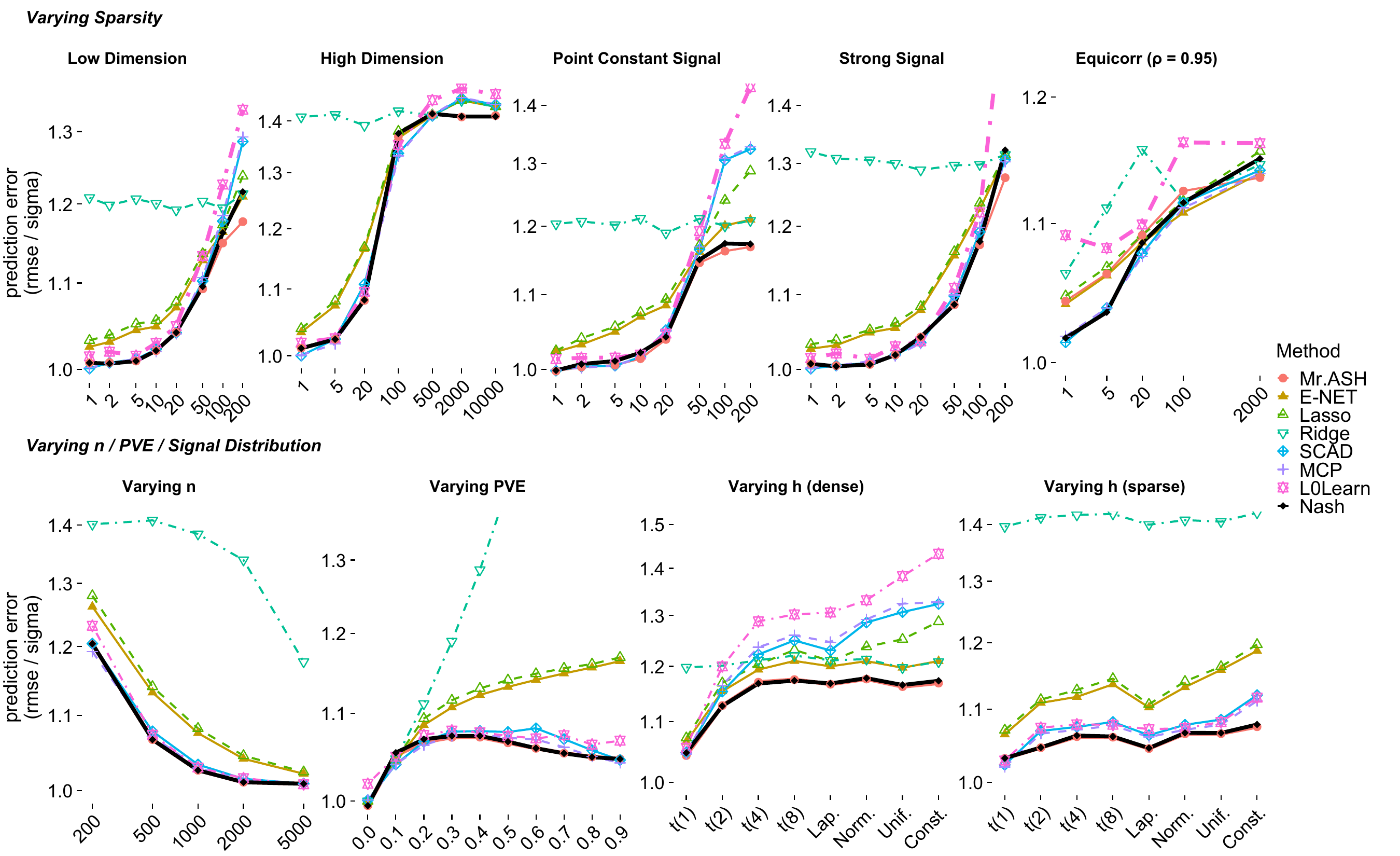}
\caption{Prediction accuracy of Nash against competing penalized regression
methods (Mr.ASH, Elastic Net, Lasso, Ridge, SCAD, MCP, L0Learn) across
simulated datasets, measured by scaled RMSE (lower is better; 1 = null
predictor). \textbf{Top row}: varying sparsity $s$ across five settings
differing in dimension, signal strength, signal distribution, and predictor
correlation. \textbf{Bottom row}: varying sample size $n$, signal-to-noise
ratio (PVE), and signal distribution $h$ in dense and sparse regimes.}
    \label{fig:placeholder1}
\end{figure}

\newpage

\subsection{Synthetic experiments with side information}\label{app:synth_sideinfo}

\paragraph{Setup:} We designed two controlled synthetic experiments to isolate the benefit of incorporating side information, holding all other factors constant. We simulate the design matrix $\mathbf{X} \in \mathbb{R}^{n \times p}$ sampling the entries from iid $N(0,1)$ with $n=500$,  and $p \in 100,500, 1\,000, 10\,000$. In both settings we compared mr.ash, Lasso, elastic-net, group-Lasso, ipf-Lasso and Nash. All the simulations were done using the following linear regression model $
  \mathbf{y} = \mathbf{X}\boldsymbol{\beta} + \boldsymbol{\varepsilon} $ where $    \boldsymbol{\varepsilon} \sim \mathcal{N}(\mathbf{0},   \mathbf{I}_n) $.

\paragraph{Known groups :} We  partitioned predictors
into five equal groups. Each group is assigned a distinct combination of sparsity and signal amplitude, ranging from a fully null group to a dense,
high-amplitude group. For each group, the effect size $\beta_j$ are  sampled from a mixture of the form $\pi_0\delta_0 +(1-\pi_0)N(0, \sigma_{s}^2)$, see   Table~\ref{tab:groups} for the simulation parameters. The side information $\mathbf{d}_j \in \{0,1\}^5$ is the one-hot
encoding of the group membership of predictor $j$.  
\begin{table}[h]
\centering
\small
\begin{tabular}{lccc}
\toprule
\textbf{Group} & \textbf{Sparsity} (prop.\  zero, $\pi_0$) & \textbf{Signal SD} $\sigma_{s}$ & \textbf{Effect type} \\
\midrule
1 & 1.00 & — & null \\
2 & 0.90 & 0.5 & weak, sparse \\
3 & 0.50 & 1.0 & medium \\
4 & 0.00 & 2.0 & dense, strong \\
5 & 0.80 & 3.0 & sparse, strong \\
\bottomrule
\end{tabular}
\caption{Simulations parameters for the \textbf{known groups} simualtions. }
\label{tab:groups}
\end{table}

\paragraph{Continuous index:} We generate $p=100;500; 1\,000, 10\,000$ predictors, each assigned a
position $t_j \sim \mathrm{Unif}(0, 3)$ along a one-dimensional index. The true
coefficient follows a spatially varying spike-and-slab: $\beta_j = 0$ within two null
bands ($t_j \in (0,0.5)$ and $t_j \in (1.5, 2.0)$), and $\beta_j \sim N(0,
\sigma(t_j)^2)$ elsewhere, where $\sigma(t_j) = 0.01 + |\sin(\pi t_j)|$ smoothly
modulates signal amplitude, we display the effect distribution along the covariate in Figure \ref{fig:illustration} panel A. The side information is $\mathbf{d}_j =  t_j  $.  In this simulation setting, group-Lasso and
ipf-Lasso, require a hand-crafted group indicator, whereas
Nash  uses only the continuous index $t$ as side information.

\paragraph{Ablation: } In both settings, we compare four 
variants of Nash: (i)~Nash-mdn (full MLP prior of the form \ref{Nashmdn}), (ii)~Nash-linear, prior of the form \ref{Nashmdn}  in which 
$\pi(\mathbf{d}_j, \boldsymbol{\theta})$ is a single affine 
layer, equivalent to a separate  ash  prior 
\citep{stephens_false_2017} per group with no nonlinear interactions, 
(iii)~Nash-no-cov (no side information), and (iv)~Nash-shuffled, where group 
labels are randomly permuted to serve as a negative control.

\begin{table}[h]
\centering
\small
\begin{tabular}{l| lcccc}
\toprule
\textbf{Side info } &\textbf{Method} & $p=100$ & $p=500$ & $p=1{,}000$ & $p=10{,}000$ \\
\midrule
Yes&group-Lasso & $0.0205\pm4e-4$ &$  0.0587\pm2e-3$ &$0.395\pm5e-3$  &  $0.472\pm2e-2$ \\
Yes&ipf-Lasso & $0.0211\pm 3 e-4$ &  $0.0601\pm1e-3$  & $0.375\pm4e-3$  & $0.453\pm2e-2$  \\
Yes&Nash-mdn (full)&$ 0.0198 \pm  5e-4$  & $ 0.0412    \pm 8 e-3$  &$0.321\pm5e-3$  &$0.407\pm6e-2$  \\
Yes&Nash-linear &$ 0.0210 \pm  3e-4$  & $ 0.0438  \pm 8 e-3$  &$0.335\pm5e-3$  &$0.418\pm8e-3$   \\
\midrule
No&Lasso & $0.0256\pm4e-4$ &$0.131\pm4e-3$  &$0.472\pm6e-2$  & $0.559\pm1e-2$  \\
No&elastic-net & $0.0228\pm3e-4$ &$0.130\pm3e-3$  &$0.469\pm8e-2$  & $0.573\pm3e-2$  \\
No&mr.ash &$ 0.0206 \pm  6e-4$  & $ 0.136   \pm 1 e-2$  &$0.429\pm6e-2$  & $0.513\pm2e-2$  \\

\midrule
No &Nash-no-cov &$ 0.0213\pm  7e-4$  & $ 0.129   \pm 9 e-3$  &$0.459\pm7e-2$  & $0.509\pm3e-2$  \\

Yes$^*$  &Nash-shuffled   &$ 0.0218\pm  7e-4$  & $ 0.135   \pm 9 e-3$  &$0.473\pm8e-2$  & $0.519\pm3e-2$  \\
\bottomrule
\end{tabular}
\caption{Simulation results, for\textbf{ the known-groups }experiment (RMSE; lower is better). $ ^*$ uninformative side information,  information, $\pm$ corresponds to the confidence interval upper and lower bounds }
\label{tab:ablation_groups}
\end{table}

\begin{table}[h]
\centering
\small
\begin{tabular}{l| lcccc}
\toprule
\textbf{Side info } &\textbf{Method} & $p=100$ & $p=500$ & $p=1{,}000$ & $p=10{,}000$ \\
\midrule
Yes$^\dagger $ &group-Lasso & $0.0225\pm4e-4$ &$  0.0562\pm2e-3$ &$0.475\pm5e-3$  &  $0.524\pm5e-3$ \\
Yes$^\dagger$ &ipf-Lasso & $0.0204\pm 4e-4 $ &  $0.0573\pm2e-3$  & $0.395\pm8e-3$ & $0.492\pm8e-3$ \\
Yes&Nash-mdn (full)&$ 0.0191 \pm  6e-4$  & $ 0.0512   \pm 8 e-3$  &$ 0.403   \pm 7e-3$   &$0.482\pm8e-3$\\
Yes&Nash-linear &$ 0.0213 \pm  5e-4$   & $ 0.131   \pm 1 e-2$  &$0.482\pm8e-2$  &  $0.534\pm5e-2$    \\
\midrule
No&Lasso & $0.0235\pm4e-4$ &$0.130\pm4e-3$  &$0.478\pm7e-2$  &$0.579\pm2e-2$   \\
No&elastic-net & $0.0225\pm4e-4$ &$0.129\pm4e-3$  &$0.459\pm7e-2$  &$0.552\pm2e-2$  \\
No&mr.ash &$ 0.0206 \pm  6e-4$  & $ 0.136   \pm 1 e-2$  &$0.439\pm7e-3$ &$0.512\pm1e-2$  \\

\midrule
No &Nash-no-cov &$ 0.0183 \pm  5e-4$  & $ 0.129   \pm 7 e-3$  &$0.429\pm8e-3$ &$0.503\pm2e-2$  \\

Yes$^*$  &Nash-shuffled &$ 0.0193 \pm  6e-4$  & $ 0.132   \pm 8 e-3$  &$0.483\pm7e-2$  &  $0.532\pm5e-2$       \\
\bottomrule
\end{tabular}
\caption{Simulation results, for the \textbf{continuous index} experiment (RMSE; lower is better).$ ^*$ uninformative side information, $^\dagger $ ipf Lasso and group Lasso cannot process the underlying side information we provided the underlying group defined in our simulation (oracle like simulation),   information, $\pm$ corresponds to the confidence interval upper and lower bounds}
\label{tab:ablation_c}
\end{table}

The results show that incorporating correct side information substantially improves
performance (Nash-mdn vs Nash-no-cov). Randomly permuting the group labels
(Nash-shuffled) removes this benefit, confirming that Nash leverages meaningful
structure in the covariates rather than overfitting noise. Furthermore, replacing the
neural network with a linear mapping (Nash-linear) leads to a consistent degradation in
performance, indicating that nonlinear modeling of the side information is beneficial
even in this structured setting.

In this setting, the signal structure varies nonlinearly with $t_j$, making it
difficult to capture with linear mappings. Consistent with this, Nash outperforms
Nash-linear, highlighting the importance of flexible neural parameterization. As in the
grouped setting, shuffling the covariates eliminates the gains, confirming that the
model exploits the spatial structure encoded in $\mathbf{d}_j$.

\subsection{Real data experiment}\label{supsec:realdata}

We evaluate Nash prediction performance on 4 real data sets that have side information. 
In each of these datasets, we also benchmark the performance of mr.ash, Lasso, the elastic net (Enet) with $\alpha=0.5$, ridge regression, and when the data display group/hierarchical side information, we also benchmark the ipf Lasso. We benchmark 2 versions of Nash: i) Nash without side information and  ii)   Nash-mdn equation \ref{Nashmdn}.  We also benchmark the performance of xgboost \citep{chen_xgboost_2016},   and multi-layer perceptron (MLP) with L2 regularization. 

We selected a range of real datasets that spans from small data set  and with a limited number of covariates (e.g. Air Passenger data ) to larger data set scale such as epigenetic age prediction with nearly 500,000 predictors. 
We detail these datasets and the preprocessing in \ref{supsec:realdata}.  
In most of our experiments, we proceed as follows: we remove at random 20\% of the data for testing purposes, run the different methods on the remaining data, and evaluate the performance of each method in terms of root mean squared error (RMSE).

\begin{itemize}
    \item \textbf{SNP500}: we used daily return from of the \textbf{AAPL} symbols from SNP500 using other assets daily return. For each asset used in the predictor, we obtained the type of industry in which this asset is part of (e.g., Technology, Communication Services, Healthcare, Equity Funds ) and used it as side information.
   \item \textbf{Airpassenger}: we added noise to the Airpassenger data set and used the measurement time point as side information. 
   \item \textbf{GSE40279}: we used methylation data from individuals measurement ($p=489, 503$) to predict the age of the subject \citep{horvath_dna_2018}. We use methylation probe annotation as side information.
   \item \textbf{TCGA}: we predict individual normalized  \textit{BRCA1}  gene expression  (an important gene in breast cancer) expression level using the other genes. We use gene pathways from KEGG pathway as side information.

\end{itemize} 

\begin{table}[h]
\centering\small
\begin{tabular}{lllll}
\toprule
Dataset & $n \times p$ & Response & Side information & Type \\
\midrule
SNP500      & $235 \times 85$       & AAPL daily return      & Industry sector   & Group \\
Airpassenger & $144 \times 144$     & Passenger count        & Time index        & Continuous \\
GSE40279    & $679 \times 489{,}503$ & Chronological age     & Probe annotation  & Group \\
TCGA        & $1{,}212 \times 18{,}300$ & \textit{BRCA1} expression & KEGG pathway & Group \\
\bottomrule
\end{tabular}
\caption{Real data experiments. All datasets use an 80/20 train/test split.}\label{table:datset_desc}
\end{table}

 \subsubsection{Group-Based and Hierarchical Side Information}
\label{ssec:gp_penalty}

This section clarifies the construction of $(X,y)$, the definition of the side
information $d_j$, the neural architecture used to parameterize the prior, and how the prior induces covariate-specific shrinkage.  
This applies to the datasets \textit{SNP500}, \textit{GSE40279}, and \textit{TCGA}.

\paragraph{Side information.}
In group-structured regression, each covariate $j$ belongs to a group
$k(j)\in\{1,\dots,K\}$.  
We encode this using a one-hot vector.
\[
d_j \in \{0,1\}^K,
\qquad 
d_{j,k} = 1 \iff j \in \text{group } k.
\]
This representation allows the prior on coefficient $b_j$ to depend explicitly on the
group membership of covariate $j$.

\paragraph{Prior parameterization.}
We use a mixture density network (MDN) to produce the group-specific shrinkage prior
\[
g(b_j \mid d_j,\theta)
=
\pi_0(d_j)\,\delta_0 \;+\;
\sum_{k=1}^{M} \pi_k(d_j)\,
N\!\left(\mu_k(d_j),\, \sigma_k^2(d_j)\right),
\]
with $g_0=\delta_0$ fixed.  
The MDN learns the smooth mapping $d_j\mapsto \{\pi_k(d_j),\mu_k(d_j),\sigma_k^2(d_j)\}$
so that different groups receive different shrinkage patterns.

\paragraph{Neural network architecture.}
The MDN is implemented as a fully connected feed-forward network with three affine
layers and ReLU activations.  
Given $d_j\in\mathbb{R}^q$, the network computes
\[
h_1 = \mathrm{ReLU}(W_1 d_j + b_1), \qquad
h_2 = \mathrm{ReLU}(W_2 h_1 + b_2),
\]
\[
\pi(d_j;\theta)
= \mathrm{Softmax}(W_3 h_2 + b_3).
\]
Here,
\[
W_1 \in \mathbb{R}^{H\times q},\quad 
W_2 \in \mathbb{R}^{H\times H},\quad
W_3 \in \mathbb{R}^{(M+1)\times H},
\]
and we use $H=32$ in all experiments.  
The final layer outputs the $(M+1)$ mixture weights for the spike-and-slab prior, along
with $2M$ linear outputs for the means and log-variances of the continuous mixture
components.

All networks in this section are trained using the Adam optimizer with a learning rate $10^{-3}$ for $100$ epochs.

% ------------------------------------------------------------------------------

\subsubsection{Continuous Side Information}

In the \textit{AirPassengers} application, the side information associated with each
coefficient is continuous rather than categorical.  This section clarifies how the
side information $d_j$ is constructed and how the MDN adapts shrinkage continuously
across covariates.

\paragraph{Side information.}
When covariates are associated with continuous metadata—such as time, spatial
location, or continuous annotations—we encode each covariate $j$ using a real-valued
feature vector
\[
d_j \in \mathbb{R}^q,
\]
where $q$ depends on the application.
Examples include  
(i)~$d_j = t_j$ for time-indexed covariates,  
(ii)~$d_j = (x_j, y_j)$ for 2D spatial structure, and  
(iii)~multi-dimensional continuous annotations.  
This allows the prior strength and sparsity pattern to vary smoothly as a function of
$d_j$.

\paragraph{Prior parameterization via an MDN.}
The mixture density network used here is identical to the architecture described in
Section~\ref{ssec:gp_penalty}.  
It outputs the mixture weights and mixture parameters as continuous functions of
$d_j$.

All MDNs in this section are trained using Adam with a learning rate $10^{-3}$ for
$100$ epochs.

% ------------------------------------------------------------------------------
\subsubsection{Graph-Based Side Information}

In the MNIST denoising experiment, the covariates correspond to pixels on a
two-dimensional grid.  
This induces a natural graph structure that the prior can exploit to encourage
spatial smoothness while allowing for sharp boundaries and local adaptivity.
Below we describe the construction of the side information $d_j$, the graph
$G=(V,E)$, and the neural architecture used to parameterize the fused prior.

\paragraph{Side information from 2D coordinates.}
Each coefficient $b_j$ corresponds to a pixel located at $(d_{j_1}, d_{j_2})$ on an
$n\times n$ grid.  
We encode its spatial position as the normalized coordinate pair
\[
d_j = (d_{j_1}/n,\; d_{j_2}/n),
\]
so that both components lie in $[0,1]$.
This representation allows the prior to vary smoothly across both spatial
dimensions.

\paragraph{Graph construction.}
We equip the pixel grid with a 4-nearest-neighbor graph:
\[
E = \{(j,k): k \in \mathrm{Nbh}(j)\}, 
\qquad
\mathrm{Nbh}(j) = \{\text{up},\text{down},\text{left},\text{right}\}.
\]
This graph encodes local spatial adjacency and enables the use of message passing
to incorporate information from neighboring pixels.

\paragraph{Fused-Laplace prior.}
To encourage piecewise smoothness, we employ a fused-Laplace shrinkage prior of the
form
\begin{equation} 
    g_{\mathrm{fused}}(d_j)
    \;=\;
    z\, L(0, s_1)\;
    L\!\left(v_1(d_j),\, s_2(d_j)\right),
\end{equation}
where $L(\mu,s)$ denotes a Laplace distribution with location $\mu$ and scale $s$,
and $z$ is a normalizing constant.
The first factor shrinks $b_j$ toward zero, while the second penalizes the local
difference $v_1(d_j)$ between $b_j$ and its neighbors.  
Learning $(s_1, s_2(d_j), v_1(d_j))$ produces a data-adaptive analogue of the fused
lasso.

\paragraph{Neural parameterization via a 2-layer message-passing GNN.}
The parameters of the fused prior
\[
(s_1,\; v_1(d_j),\; s_2(d_j))
\]
are produced by a \emph{two-layer message-passing neural network} (MPNN) operating on
the graph $G=(V,E)$.
Let $h_j^{(0)} = d_j$ denote the initial node features.  
The GNN performs two rounds of message passing:
\[
h_j^{(1)} 
= 
\mathrm{ReLU}\!\left(
W_1 h_j^{(0)} 
\;+\;
\sum_{k \in \mathrm{Nbh}(j)}
    W_{\mathrm{msg}}\, h_k^{(0)}
\right),
\]
\[
h_j^{(2)} 
= 
\mathrm{ReLU}\!\left(
W_2 h_j^{(1)} 
\;+\;
\sum_{k \in \mathrm{Nbh}(j)}
    W_{\mathrm{msg}}'\, h_k^{(1)}
\right).
\]
A final linear layer produces the prior parameters:
\[
(s_1,\; s_2(d_j),\; v_1(d_j))
=
\mathrm{Softplus}\!\left(
W_3 h_j^{(2)} + b_3
\right).
\]
This architecture allows the shrinkage strength $s_1$, the smoothness scale
$s_2(d_j)$, and the fused-lasso direction $v_1(d_j)$ to depend not only on the
spatial coordinates of pixel $j$, but also on the features of its neighbors through
message passing.

We use  $H = 64$  hidden dimensions and train all GNNs using the Adam optimizer with
learning rate $10^{-3}$ for $100$ epochs.

\subsubsection{Other methods}
 
\paragraph{MLP/Feed forward neural network:}
We trained a feed-forward neural network (NN) as a baseline for comparison. The network consists of three fully connected layers: an input layer followed by two hidden layers of sizes 128 and 64, each with ReLU activation. The output is a single linear unit predicting a continuous response.  The input features were standardized to zero mean and unit variance based on the training set. The model was trained using the Adam optimizer with a learning rate of 0.001 and mean squared error (MSE) loss. Each model was trained for 100 epochs using all training samples in batch mode. Training and evaluation were implemented in PyTorch. RMSE was computed on each test split, and average RMSE across the 10 folds is reported.

\paragraph{xgboost:} 

We trained gradient-boosted decision trees using the xgboost \textit{R} package with default hyperparameters. Specifically, we used root mean square error as objective for regression. Each model was trained using 50 boosting iterations (default parameter) with a maximum tree depth of 6 and a learning rate (eta) of 0.3.

 \paragraph{MNIST}
We evaluated seven denoising methods on the MNIST dataset with additive Gaussian noise of standard deviation $\sigma = 0.2$, applied independently to each pixel. Each method was applied to 20 randomly selected test images, and performance was measured using root mean squared error (RMSE) against the clean image. The Noise2Self model was a convolutional neural network (CNN) with three convolutional layers: $\text{Conv}_{1\to32} \to \text{ReLU} \to \text{Conv}_{32\to32} \to \text{ReLU} \to \text{Conv}_{32\to1}$. It was trained using masked pixel regression, where approximately 10\% of pixels were randomly set to zero during each training iteration and the model was trained to reconstruct them. The loss was computed only over masked pixels using mean squared error. We trained the Noise2Self model for 5 epochs using the Adam optimizer with a learning rate of $10^{-3}$ and batch size 64.

The Nash-fused was trained using a message passing GNN representing each image as a 784-node graph corresponding to a $28 \times 28$ grid, with 4-neighbor connectivity and node features consisting of the noisy intensity and normalized spatial coordinates. The underlying graph neural network had two hidden layers with ReLU activations. We trained Nash-fused separately for each image using 300 steps of gradient descent with the Adam optimizer and learning rate $10^{-2}$.

Classical baselines included total variation (TV) denoising, fused lasso, Gaussian smoothing, and non-local means (NLM). TV denoising used  regularization weight 0.1. Fused lasso was formulated as a convex optimization problem with an $\ell_2$ data fidelity term and isotropic TV penalty, solved using \texttt{cvxpy} with the SCS solver. Gaussian filtering used a fixed kernel with standard deviation $\sigma=1$. NLM used the implementation from \texttt{skimage.restoration} with parameters $h = 1.15 \cdot \hat{\sigma}$ (where $\hat{\sigma}$ is estimated from the image), patch size 3, and patch distance 5.  
 
%This serves as a strong non-Bayesian neural baseline for denoising.

\begin{figure}
    \centering
    \includegraphics[width=0.9\linewidth]{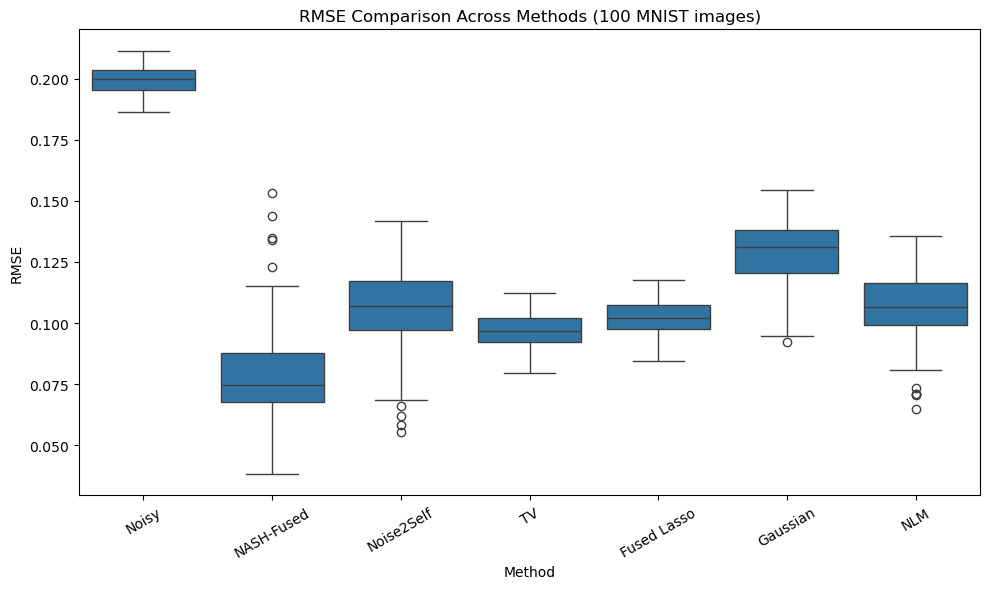}
    \caption{ performances of the different approaches for denoising 100 MNIST images in terms of RMSE.}
    \label{fig:cnn}
\end{figure}

\begin{figure}
    \centering
    \includegraphics[width=0.9\linewidth]{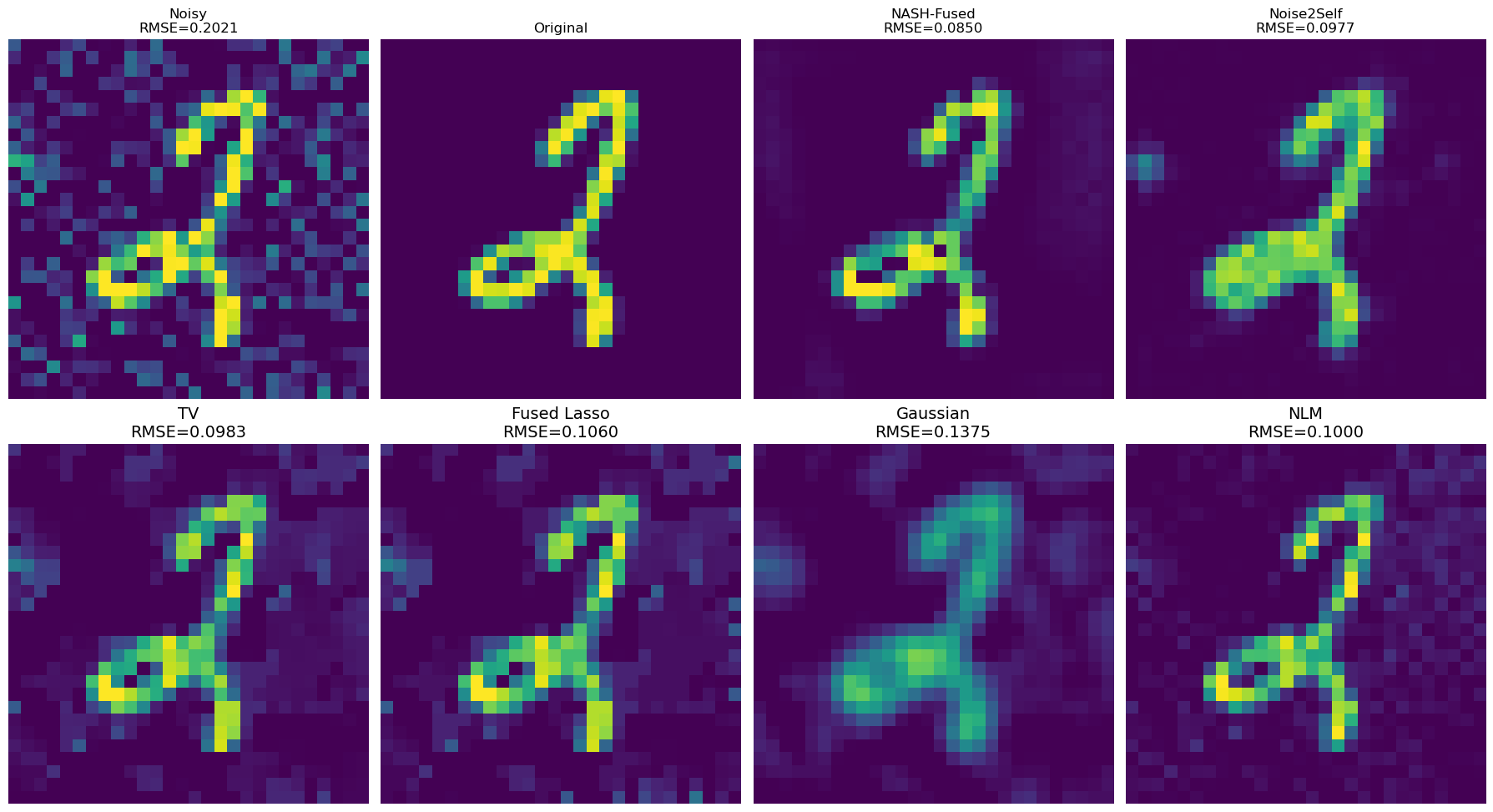}
     \caption{Additional denoised image}
    \label{fig:mnist2}
\end{figure}

\begin{figure}
    \centering
    \includegraphics[width=0.9\linewidth]{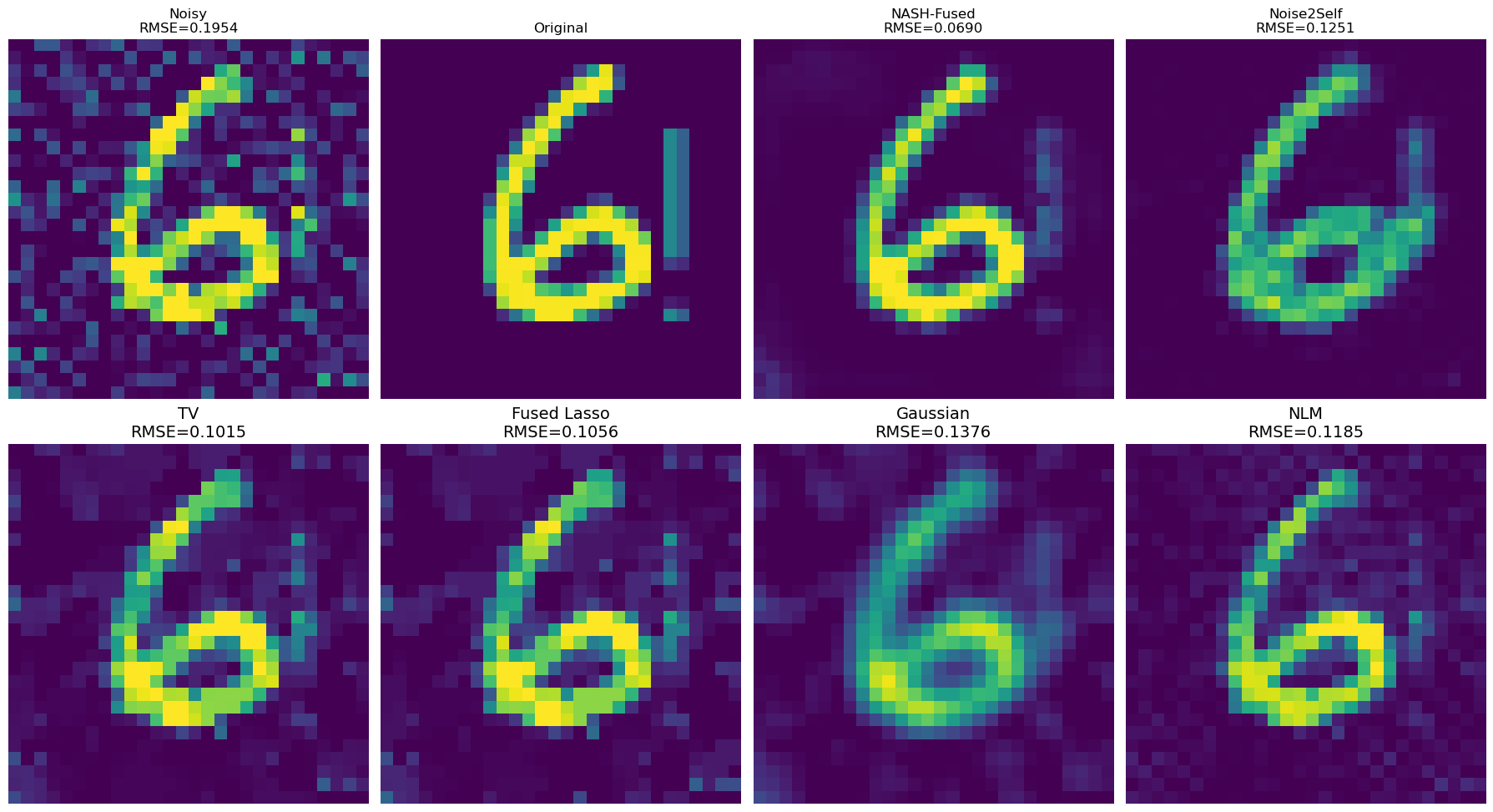}
     \caption{Additional denoised image}
    \label{fig:mnist3}
\end{figure}

\begin{figure}
    \centering
    \includegraphics[width=0.9\linewidth]{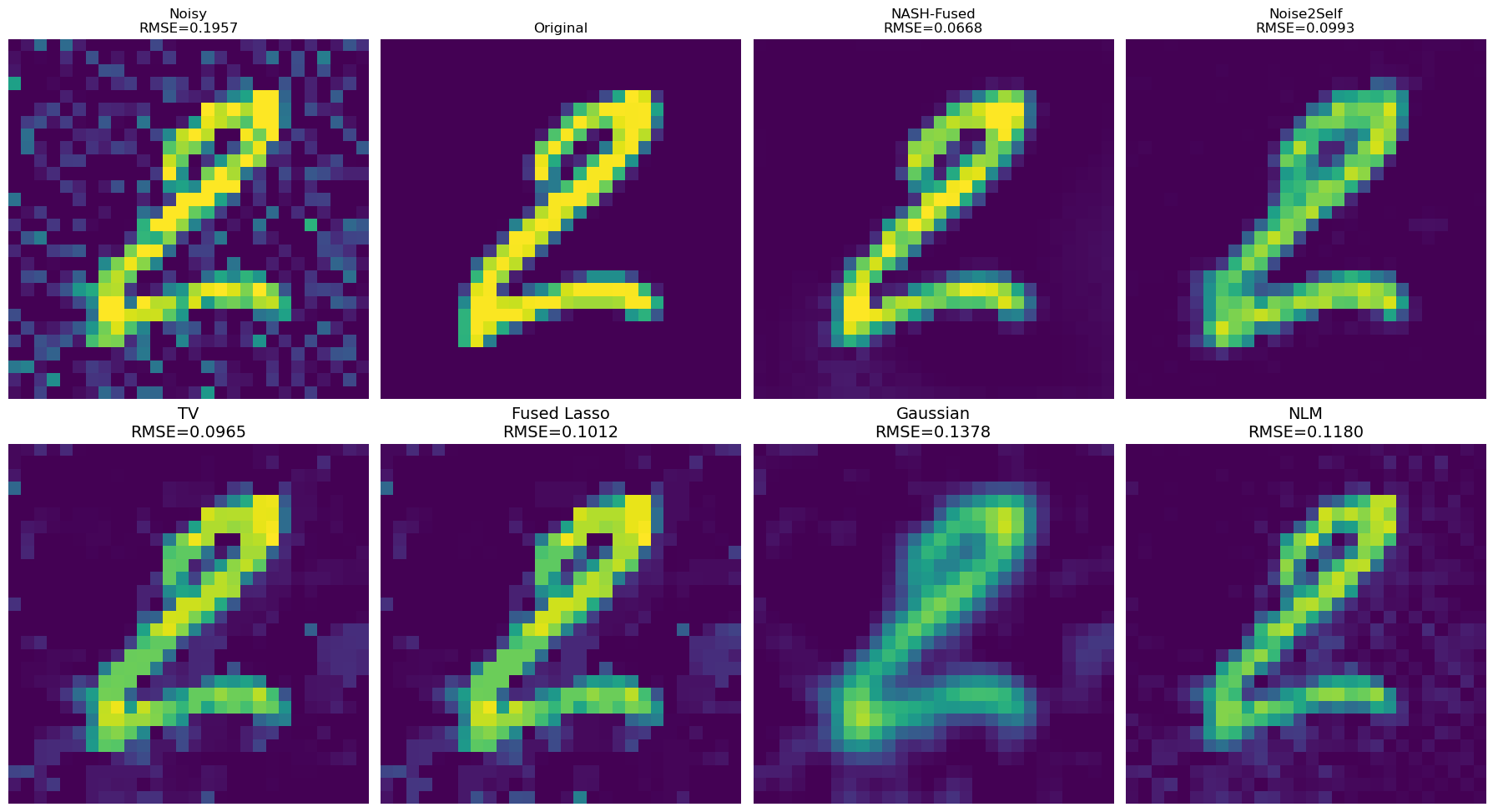}
     \caption{Additional denoised image}
    \label{fig:mnist4}
\end{figure}

\begin{figure}
    \centering
    \includegraphics[width=0.9\linewidth]{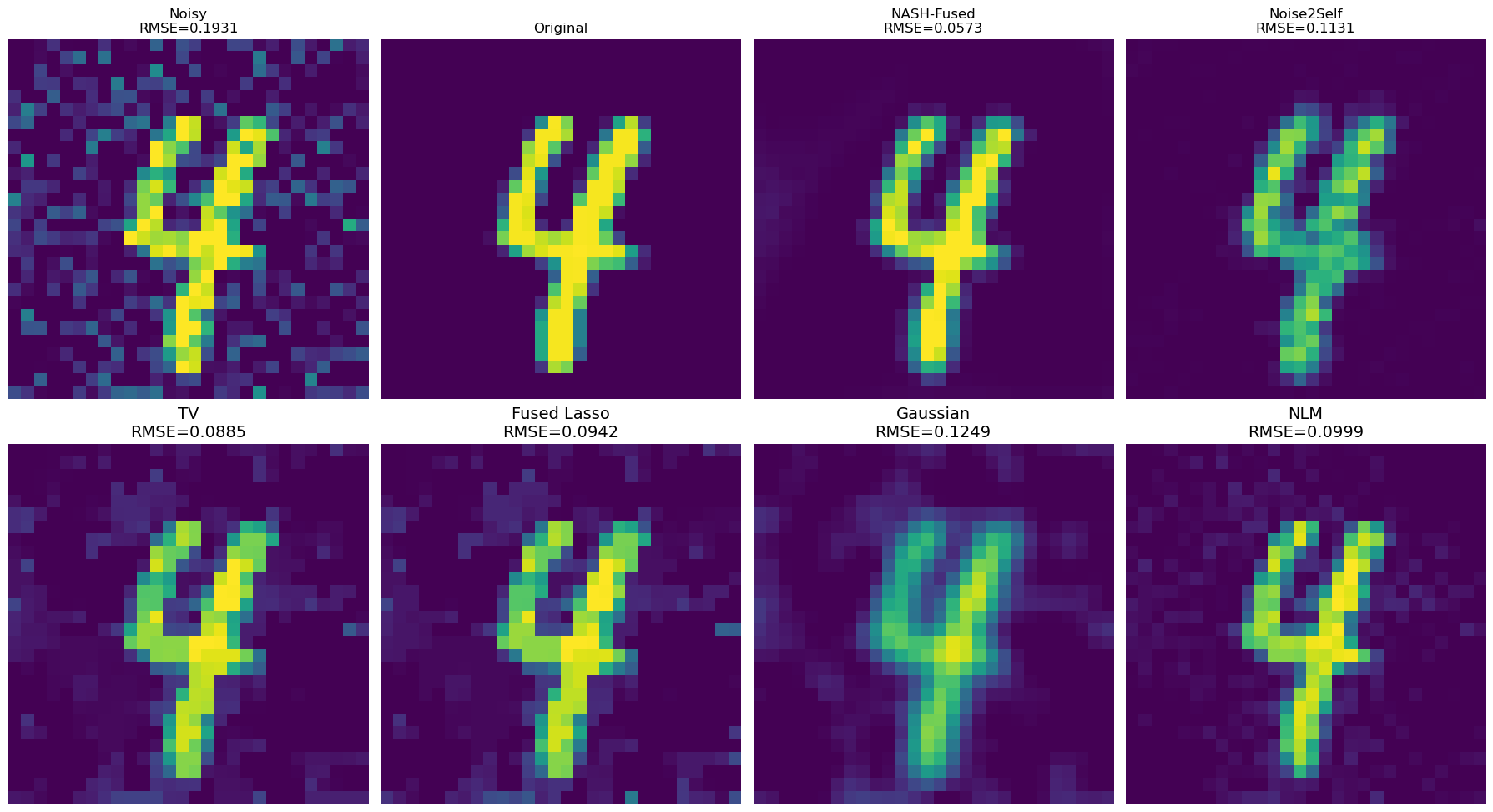}
     \caption{Additional denoised image}
    \label{fig:mnist5}
\end{figure}
\begin{figure}
    \centering
    \includegraphics[width=0.9\linewidth]{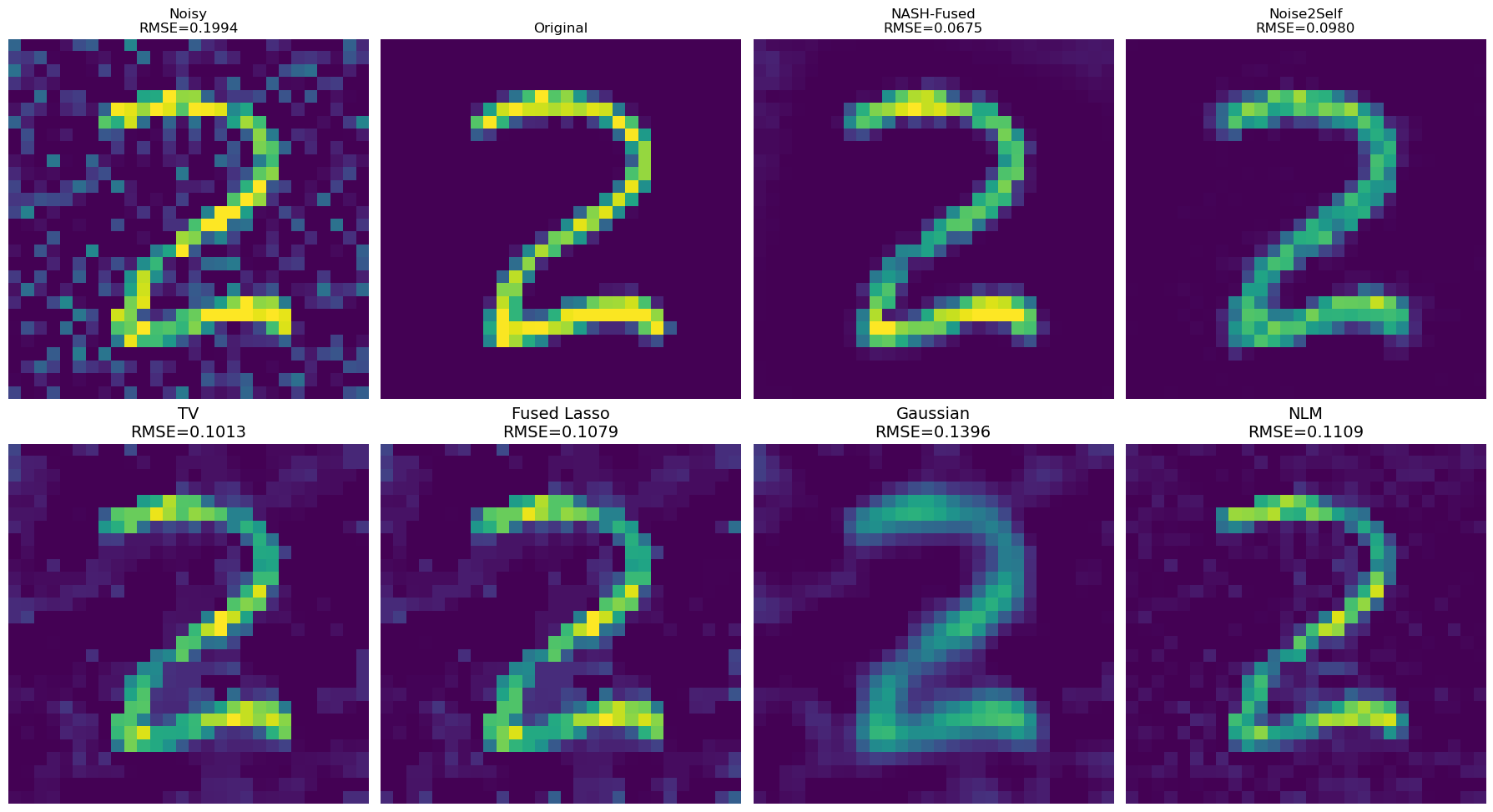}
     \caption{Additional denoised image}
    \label{fig:mnist6}
\end{figure}\begin{figure}
    \centering
    \includegraphics[width=0.9\linewidth]{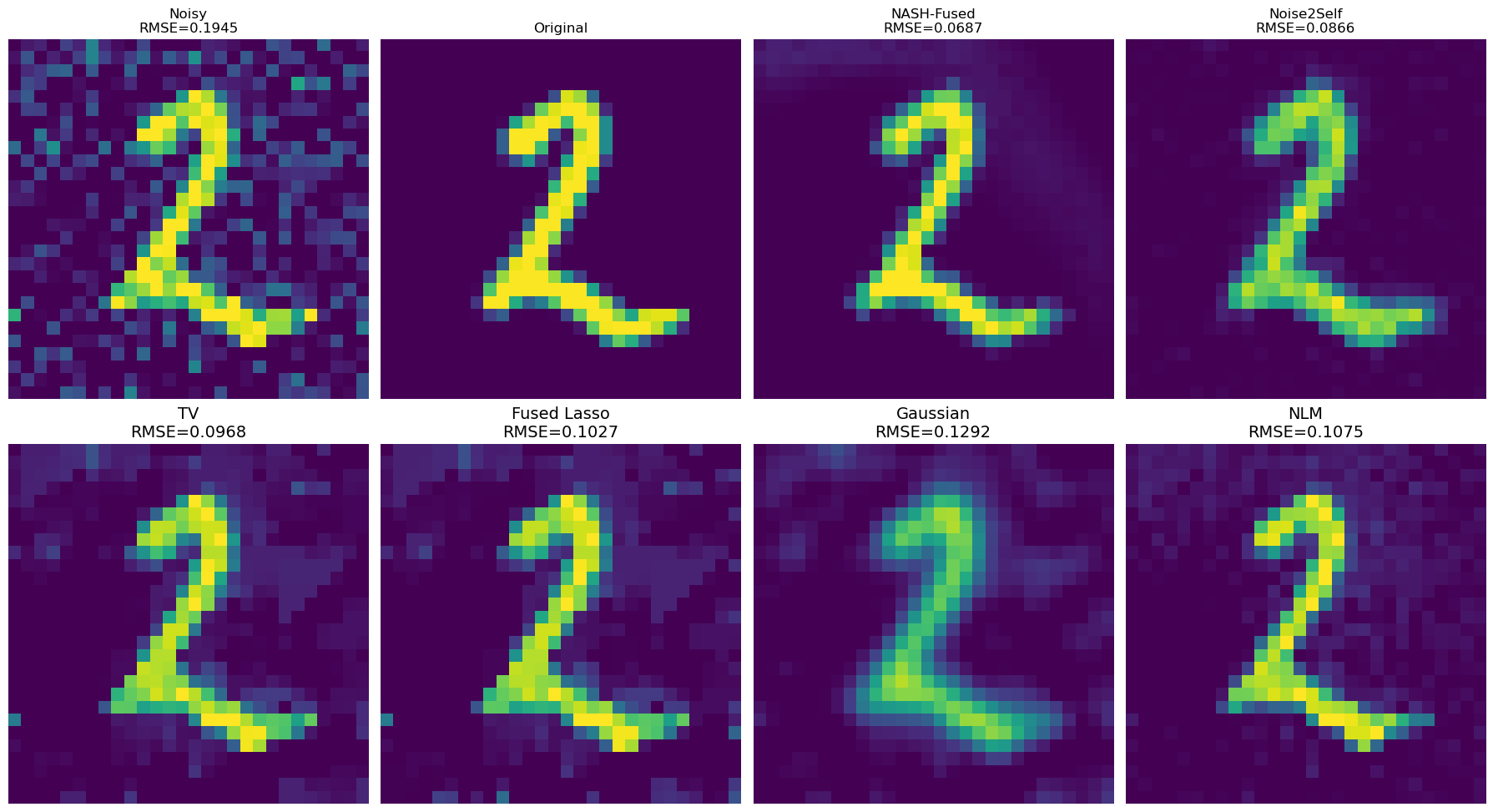}
     \caption{Additional denoised image}
    \label{fig:mnist7}
\end{figure}\begin{figure}
    \centering
    \includegraphics[width=0.9\linewidth]{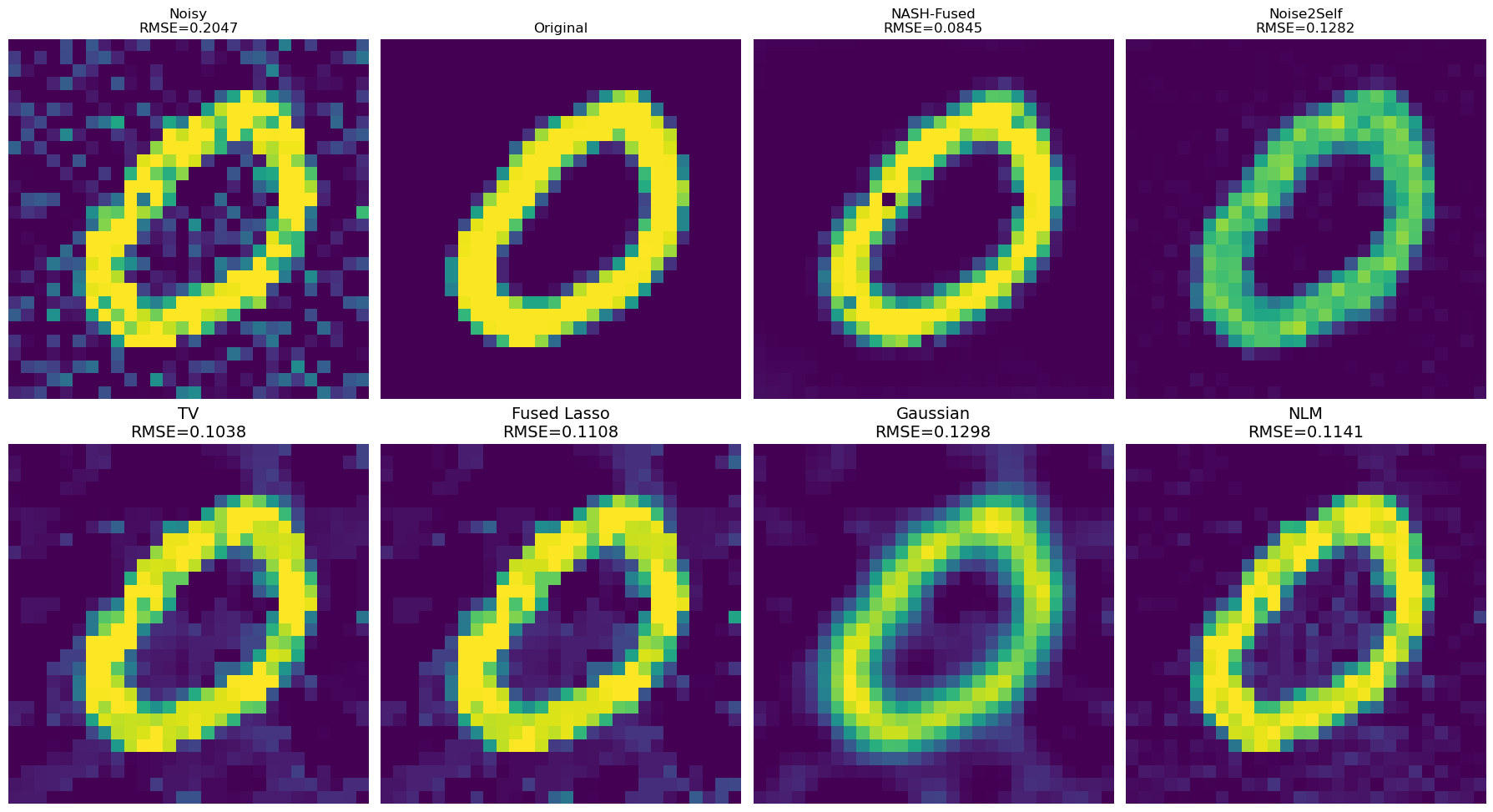}
     \caption{Additional denoised image}
    \label{fig:mnist8}
\end{figure}

\newpage

%% file: library.bib
@article{stephens_false_2017,
	title = {False discovery rates: a new deal},
	volume = {18},
	issn = {1465-4644},
	shorttitle = {False discovery rates},
	url = {https://academic.oup.com/biostatistics/article/18/2/275/2557030},
	doi = {10.1093/biostatistics/kxw041},
	language = {en},
	number = {2},
	journal = {Biostatistics},
	author = {Stephens, Matthew},
	month = apr,
	year = {2017},
	pages = {275--294},
}

@article{horvath_dna_2018,
	title = {{DNA} methylation-based biomarkers and the epigenetic clock theory of ageing},
	volume = {19},
	copyright = {2018 Macmillan Publishers Ltd., part of Springer Nature},
	issn = {1471-0064},
	url = {https://www.nature.com/articles/s41576-018-0004-3},
	doi = {10.1038/s41576-018-0004-3},
	language = {En},
	number = {6},
	urldate = {2019-06-27},
	journal = {Nature Reviews Genetics},
	author = {Horvath, Steve and Raj, Kenneth},
	month = jun,
	year = {2018},
	pages = {371},
}

@article{tibshirani_sparsity_2005,
	title = {Sparsity and smoothness via the fused lasso},
	volume = {67},
	issn = {1467-9868},
	url = {https://rss.onlinelibrary.wiley.com/doi/abs/10.1111/j.1467-9868.2005.00490.x},
	doi = {10.1111/j.1467-9868.2005.00490.x},
	language = {en},
	number = {1},
	urldate = {2020-02-11},
	journal = {Journal of the Royal Statistical Society: Series B (Statistical Methodology)},
	author = {Tibshirani, Robert and Saunders, Michael and Rosset, Saharon and Zhu, Ji and Knight, Keith},
	year = {2005},
	pages = {91--108},
}

@article{tibshirani_regression_1996,
	title = {Regression {Shrinkage} and {Selection} {Via} the {Lasso}},
	volume = {58},
	copyright = {© 1996 Royal Statistical Society},
	issn = {2517-6161},
	url = {https://rss.onlinelibrary.wiley.com/doi/abs/10.1111/j.2517-6161.1996.tb02080.x},
	doi = {10.1111/j.2517-6161.1996.tb02080.x},
	language = {en},
	number = {1},
	urldate = {2020-02-11},
	journal = {Journal of the Royal Statistical Society: Series B (Methodological)},
	author = {Tibshirani, Robert},
	year = {1996},
	pages = {267--288},
}

@inproceedings{chen_xgboost_2016,
	address = {San Francisco California USA},
	title = {{XGBoost}: {A} {Scalable} {Tree} {Boosting} {System}},
	isbn = {978-1-4503-4232-2},
	shorttitle = {{XGBoost}},
	url = {https://dl.acm.org/doi/10.1145/2939672.2939785},
	doi = {10.1145/2939672.2939785},
	language = {en},
	urldate = {2020-10-11},
	booktitle = {Proceedings of the 22nd {ACM} {SIGKDD} {International} {Conference} on {Knowledge} {Discovery} and {Data} {Mining}},
	publisher = {ACM},
	author = {Chen, Tianqi and Guestrin, Carlos},
	month = aug,
	year = {2016},
	pages = {785--794},
}

@article{blei_variational_2017,
	title = {Variational {Inference}: {A} {Review} for {Statisticians}},
	volume = {112},
	issn = {0162-1459},
	shorttitle = {Variational {Inference}},
	url = {https://doi.org/10.1080/01621459.2017.1285773},
	doi = {10.1080/01621459.2017.1285773},
	number = {518},
	urldate = {2022-04-15},
	journal = {Journal of the American Statistical Association},
	publisher = {Taylor \& Francis},
	author = {Blei, David M. and Kucukelbir, Alp and McAuliffe, Jon D.},
	month = apr,
	year = {2017},
	note = {\_eprint: https://doi.org/10.1080/01621459.2017.1285773},
	pages = {859--877},
}

@article{wang_empirical_2021,
	title = {Empirical {Bayes} {Matrix} {Factorization}},
	volume = {22},
	issn = {1533-7928},
	url = {http://jmlr.org/papers/v22/20-589.html},
	number = {120},
	urldate = {2022-08-17},
	journal = {Journal of Machine Learning Research},
	author = {Wang, Wei and Stephens, Matthew},
	year = {2021},
	pages = {1--40},
}

@phdthesis{xie_empirical_2023,
	title = {Empirical {Bayes} methods for count data.},
	author = {Xie, Dongyue},
	year = {2023},
}

@article{kipf_semi-supervised_2017,
	title = {{SEMI}-{SUPERVISED} {CLASSIFICATION} {WITH} {GRAPH} {CONVOLUTIONAL} {NETWORKS}},
	language = {en},
	author = {Kipf, Thomas N and Welling, Max},
	year = {2017},
}

@misc{willwerscheid_ebnm_2024,
	title = {ebnm: {An} {R} {Package} for {Solving} the {Empirical} {Bayes} {Normal} {Means} {Problem} {Using} a {Variety} of {Prior} {Families}},
	shorttitle = {ebnm},
	url = {http://arxiv.org/abs/2110.00152},
	doi = {10.48550/arXiv.2110.00152},
	urldate = {2024-05-10},
	publisher = {arXiv},
	author = {Willwerscheid, Jason and Carbonetto, Peter and Stephens, Matthew},
	month = mar,
	year = {2024},
	note = {arXiv:2110.00152 [stat]},
}

@article{devriendt_sparse_2021,
	title = {Sparse {Regression} with {Multi}-type {Regularized} {Feature} {Modeling}},
	volume = {96},
	issn = {01676687},
	url = {http://arxiv.org/abs/1810.03136},
	doi = {10.1016/j.insmatheco.2020.11.010},
	language = {en},
	urldate = {2025-03-21},
	journal = {Insurance: Mathematics and Economics},
	author = {Devriendt, Sander and Antonio, Katrien and Reynkens, Tom and Verbelen, Roel},
	month = jan,
	year = {2021},
	note = {arXiv:1810.03136 [stat]},
	pages = {248--261},
}

@article{boulesteix_ipf-lasso_2017,
	title = {{IPF}-{LASSO}: {Integrative} {L1}-{Penalized} {Regression} with {Penalty} {Factors} for {Prediction} {Based} on {Multi}-{Omics} {Data}},
	volume = {2017},
	issn = {1748-6718},
	shorttitle = {{IPF}-{LASSO}},
	doi = {10.1155/2017/7691937},
	language = {eng},
	journal = {Comput Math Methods Med},
	author = {Boulesteix, Anne-Laure and De Bin, Riccardo and Jiang, Xiaoyu and Fuchs, Mathias},
	year = {2017},
	pages = {7691937},
}

@article{yu_sparse_2016,
	title = {Sparse {Regression} {Incorporating} {Graphical} {Structure} {Among} {Predictors}},
	volume = {111},
	issn = {0162-1459},
	url = {https://doi.org/10.1080/01621459.2015.1034319},
	doi = {10.1080/01621459.2015.1034319},
	number = {514},
	urldate = {2025-03-21},
	journal = {Journal of the American Statistical Association},
	publisher = {ASA Website},
	author = {Yu, Guan and and Liu, Yufeng},
	month = apr,
	year = {2016},
	note = {\_eprint: https://doi.org/10.1080/01621459.2015.1034319},
	pages = {707--720},
}

@article{gertheiss_sparse_2010,
	title = {Sparse modeling of categorial explanatory variables},
	volume = {4},
	issn = {1932-6157, 1941-7330},
	url = {https://projecteuclid.org/journals/annals-of-applied-statistics/volume-4/issue-4/Sparse-modeling-of-categorial-explanatory-variables/10.1214/10-AOAS355.full},
	doi = {10.1214/10-AOAS355},
	number = {4},
	urldate = {2025-03-21},
	journal = {The Annals of Applied Statistics},
	publisher = {Institute of Mathematical Statistics},
	author = {Gertheiss, Jan and Tutz, Gerhard},
	month = dec,
	year = {2010},
	pages = {2150--2180},
}

@article{tutz_modelling_2017,
	title = {Modelling {Clustered} {Heterogeneity}: {Fixed} {Effects}, {Random} {Effects} and {Mixtures}},
	volume = {85},
	copyright = {© 2016 The Authors. International Statistical Review © 2016 International Statistical Institute},
	issn = {1751-5823},
	shorttitle = {Modelling {Clustered} {Heterogeneity}},
	url = {https://onlinelibrary.wiley.com/doi/abs/10.1111/insr.12161},
	doi = {10.1111/insr.12161},
	language = {en},
	number = {2},
	urldate = {2025-03-21},
	journal = {International Statistical Review},
	author = {Tutz, Gerhard and Oelker, Margret-Ruth},
	year = {2017},
	note = {\_eprint: https://onlinelibrary.wiley.com/doi/pdf/10.1111/insr.12161},
	pages = {204--227},
}

@article{yuan_model_2006,
	title = {Model selection and estimation in regression with grouped variables},
	volume = {68},
	issn = {1467-9868},
	url = {https://onlinelibrary.wiley.com/doi/abs/10.1111/j.1467-9868.2005.00532.x},
	doi = {10.1111/j.1467-9868.2005.00532.x},
	language = {en},
	number = {1},
	urldate = {2025-03-24},
	journal = {Journal of the Royal Statistical Society: Series B (Statistical Methodology)},
	author = {Yuan, Ming and Lin, Yi},
	year = {2006},
	note = {\_eprint: https://onlinelibrary.wiley.com/doi/pdf/10.1111/j.1467-9868.2005.00532.x},
	pages = {49--67},
}

@article{tibshirani_solution_2011,
	title = {The solution path of the generalized lasso},
	volume = {39},
	issn = {0090-5364, 2168-8966},
	url = {https://projecteuclid.org/journals/annals-of-statistics/volume-39/issue-3/The-solution-path-of-the-generalized-lasso/10.1214/11-AOS878.full},
	doi = {10.1214/11-AOS878},
	number = {3},
	urldate = {2025-03-24},
	journal = {The Annals of Statistics},
	publisher = {Institute of Mathematical Statistics},
	author = {Tibshirani, Ryan J. and Taylor, Jonathan},
	month = jun,
	year = {2011},
	pages = {1335--1371},
}

@article{lemhadri_lassonet_2021,
	title = {{LassoNet}: {A} {Neural} {Network} with {Feature} {Sparsity}},
	volume = {22},
	issn = {1533-7928},
	shorttitle = {{LassoNet}},
	url = {http://jmlr.org/papers/v22/20-848.html},
	number = {127},
	urldate = {2025-03-24},
	journal = {Journal of Machine Learning Research},
	author = {Lemhadri, Ismael and Ruan, Feng and Abraham, Louis and Tibshirani, Robert},
	year = {2021},
	pages = {1--29},
}

@article{zou_regularization_2005,
	title = {Regularization and {Variable} {Selection} {Via} the {Elastic} {Net}},
	volume = {67},
	copyright = {https://academic.oup.com/journals/pages/open\_access/funder\_policies/chorus/standard\_publication\_model},
	issn = {1369-7412, 1467-9868},
	url = {https://academic.oup.com/jrsssb/article/67/2/301/7109482},
	doi = {10.1111/j.1467-9868.2005.00503.x},
	language = {en},
	number = {2},
	urldate = {2025-03-24},
	journal = {Journal of the Royal Statistical Society Series B: Statistical Methodology},
	author = {Zou, Hui and Hastie, Trevor},
	month = apr,
	year = {2005},
	pages = {301--320},
}

@inproceedings{nalisnick_hybrid_2019,
	title = {Hybrid {Models} with {Deep} and {Invertible} {Features}},
	issn = {2640-3498},
	url = {https://proceedings.mlr.press/v97/nalisnick19b.html},
	language = {en},
	urldate = {2025-03-24},
	booktitle = {Proceedings of the 36th {International} {Conference} on {Machine} {Learning}},
	publisher = {PMLR},
	author = {Nalisnick, Eric and Matsukawa, Akihiro and Teh, Yee Whye and Gorur, Dilan and Lakshminarayanan, Balaji},
	month = may,
	year = {2019},
	pages = {4723--4732},
}

@article{okoh_hybrid_2018,
	title = {A {Hybrid} {Regression}-{Neural} {Network} ({HR}-{NN}) {Method} for {Forecasting} the {Solar} {Activity}},
	volume = {16},
	copyright = {©2018. American Geophysical Union. All Rights Reserved.},
	issn = {1542-7390},
	url = {https://onlinelibrary.wiley.com/doi/abs/10.1029/2018SW001907},
	doi = {10.1029/2018SW001907},
	language = {en},
	number = {9},
	urldate = {2025-03-24},
	journal = {Space Weather},
	author = {Okoh, D. I. and Seemala, G. K. and Rabiu, A. B. and Uwamahoro, J. and Habarulema, J. B. and Aggarwal, M.},
	year = {2018},
	note = {\_eprint: https://onlinelibrary.wiley.com/doi/pdf/10.1029/2018SW001907},
	pages = {1424--1436},
}

@article{oelker_uniform_2017,
	title = {A uniform framework for the combination of penalties in generalized structured models},
	volume = {11},
	issn = {1862-5355},
	url = {https://doi.org/10.1007/s11634-015-0205-y},
	doi = {10.1007/s11634-015-0205-y},
	language = {en},
	number = {1},
	urldate = {2025-03-24},
	journal = {Adv Data Anal Classif},
	author = {Oelker, Margret-Ruth and Tutz, Gerhard},
	month = mar,
	year = {2017},
	pages = {97--120},
}

@inproceedings{errica_graph_2021,
	series = {Proceedings of {Machine} {Learning} {Research}},
	title = {Graph {Mixture} {Density} {Networks}},
	volume = {139},
	url = {https://proceedings.mlr.press/v139/errica21a.html},
	booktitle = {Proceedings of the 38th {International} {Conference} on {Machine} {Learning}},
	publisher = {PMLR},
	author = {Errica, Federico and Bacciu, Davide and Micheli, Alessio},
	year = {2021},
	pages = {3025--3035},
}

@article{polson_proximal_2015,
	title = {Proximal {Algorithms} in {Statistics} and {Machine} {Learning}},
	volume = {30},
	issn = {0883-4237, 2168-8745},
	url = {https://projecteuclid.org/journals/statistical-science/volume-30/issue-4/Proximal-Algorithms-in-Statistics-and-Machine-Learning/10.1214/15-STS530.full},
	doi = {10.1214/15-STS530},
	number = {4},
	urldate = {2025-04-01},
	journal = {Statistical Science},
	publisher = {Institute of Mathematical Statistics},
	author = {Polson, Nicholas G. and Scott, James G. and Willard, Brandon T.},
	month = nov,
	year = {2015},
	pages = {559--581},
}

@techreport{bishop_mixture_1994,
	address = {Birmingham, UK},
	type = {Technical {Report}},
	title = {Mixture {Density} {Networks}},
	url = {https://publications.aston.ac.uk/id/eprint/373/1/NCRG_94_004.pdf},
	number = {NCRG/94/004},
	institution = {Aston University},
	author = {Bishop, Christopher M.},
	year = {1994},
}

@article{hoerl_ridge_1970,
	title = {Ridge {Regression}: {Biased} {Estimation} for {Nonorthogonal} {Problems}},
	volume = {12},
	issn = {0040-1706},
	shorttitle = {Ridge {Regression}},
	url = {https://www.jstor.org/stable/1267351},
	doi = {10.2307/1267351},
	number = {1},
	urldate = {2025-04-07},
	journal = {Technometrics},
	publisher = {[Taylor \& Francis, Ltd., American Statistical Association, American Society for Quality]},
	author = {Hoerl, Arthur E. and Kennard, Robert W.},
	year = {1970},
	pages = {55--67},
}

@article{hazimeh_l0learn_2023,
	title = {{L0Learn}: {A} {Scalable} {Package} for {Sparse} {Learning} using {L0} {Regularization}},
	volume = {24},
	issn = {1533-7928},
	shorttitle = {{L0Learn}},
	url = {http://jmlr.org/papers/v24/22-0189.html},
	number = {205},
	urldate = {2025-04-07},
	journal = {Journal of Machine Learning Research},
	author = {Hazimeh, Hussein and Mazumder, Rahul and Nonet, Tim},
	year = {2023},
	pages = {1--8},
}

@article{breheny_coordinate_2011,
	title = {Coordinate descent algorithms for nonconvex penalized regression, with applications to biological feature selection},
	volume = {5},
	issn = {1932-6157, 1941-7330},
	url = {https://projecteuclid.org/journals/annals-of-applied-statistics/volume-5/issue-1/Coordinate-descent-algorithms-for-nonconvex-penalized-regression-with-applications-to/10.1214/10-AOAS388.full},
	doi = {10.1214/10-AOAS388},
	number = {1},
	urldate = {2025-04-07},
	journal = {The Annals of Applied Statistics},
	publisher = {Institute of Mathematical Statistics},
	author = {Breheny, Patrick and Huang, Jian},
	month = mar,
	year = {2011},
	pages = {232--253},
}

@article{kim_flexible_2024,
	title = {A flexible empirical {Bayes} approach to multiple linear regression and connections with penalized regression},
	volume = {25},
	issn = {1533-7928},
	url = {http://jmlr.org/papers/v25/22-0953.html},
	number = {185},
	urldate = {2025-04-09},
	journal = {Journal of Machine Learning Research},
	author = {Kim, Youngseok and Wang, Wei and Carbonetto, Peter and Stephens, Matthew},
	year = {2024},
	pages = {1--59},
}

@incollection{chipman_practical_2001,
	series = {{IMS} {Lecture} {Notes} – {Monograph} {Series}},
	title = {The {Practical} {Implementation} of {Bayesian} {Model} {Selection}},
	volume = {38},
	booktitle = {Model {Selection}},
	publisher = {Institute of Mathematical Statistics},
	author = {Chipman, Hugh and George, Edward I. and McCulloch, Robert E.},
	year = {2001},
	pages = {65--116},
}

@article{betancourt_bayesian_2017,
	title = {Bayesian {Fused} {Lasso} {Regression} for {Dynamic} {Binary} {Networks}},
	copyright = {© 2017 American Statistical Association, Institute of Mathematical Statistics, and Interface Foundation of North America},
	issn = {1061-8600},
	url = {https://www.tandfonline.com/doi/full/10.1080/10618600.2017.1341323},
	language = {EN},
	urldate = {2025-04-10},
	journal = {Journal of Computational and Graphical Statistics},
	publisher = {Taylor \& Francis},
	author = {Betancourt, Brenda and Rodríguez, Abel and Boyd, Naomi},
	month = oct,
	year = {2017},
}

@article{casella_penalized_2010,
	title = {Penalized regression, standard errors, and {Bayesian} lassos},
	volume = {5},
	issn = {1936-0975, 1931-6690},
	url = {https://projecteuclid.org/journals/bayesian-analysis/volume-5/issue-2/Penalized-regression-standard-errors-and-Bayesian-lassos/10.1214/10-BA607.full},
	doi = {10.1214/10-BA607},
	number = {2},
	urldate = {2025-04-10},
	journal = {Bayesian Analysis},
	publisher = {International Society for Bayesian Analysis},
	author = {Casella, George and Ghosh, Malay and Gill, Jeff and Kyung, Minjung},
	month = jun,
	year = {2010},
	pages = {369--411},
}

@inproceedings{buades_non-local_2005,
	title = {A non-local algorithm for image denoising},
	volume = {2},
	issn = {1063-6919},
	url = {https://ieeexplore.ieee.org/document/1467423},
	doi = {10.1109/CVPR.2005.38},
	urldate = {2025-05-15},
	booktitle = {2005 {IEEE} {Computer} {Society} {Conference} on {Computer} {Vision} and {Pattern} {Recognition} ({CVPR}'05)},
	author = {Buades, A. and Coll, B. and Morel, J.-M.},
	month = jun,
	year = {2005},
	pages = {60--65 vol. 2},
}

@article{chambolle_algorithm_2004,
	title = {An algorithm for total variation minimization and applications},
	volume = {20},
	number = {1-2},
	journal = {Journal of Mathematical Imaging and Vision},
	publisher = {Springer},
	author = {Chambolle, Antonin},
	year = {2004},
	pages = {89--97},
}

@book{gonzalez_digital_2002,
	address = {Upper Saddle River, NJ, USA},
	edition = {2nd},
	title = {Digital {Image} {Processing}},
	publisher = {Prentice Hall},
	author = {Gonzalez, Rafael C. and Woods, Richard E.},
	year = {2002},
}

@article{huang_fast_1979,
	title = {A fast two-dimensional median filtering algorithm},
	volume = {27},
	issn = {0096-3518},
	url = {https://ieeexplore.ieee.org/document/1163188},
	doi = {10.1109/TASSP.1979.1163188},
	number = {1},
	urldate = {2025-05-15},
	journal = {IEEE Transactions on Acoustics, Speech, and Signal Processing},
	author = {Huang, T. and Yang, G. and Tang, G.},
	month = feb,
	year = {1979},
	pages = {13--18},
}

@inproceedings{batson_noise2self_2019,
	series = {Proceedings of {Machine} {Learning} {Research}},
	title = {{Noise2Self}: {Blind} {Denoising} by {Self}-{Supervision}},
	volume = {97},
	booktitle = {Proceedings of the 36th {International} {Conference} on {Machine} {Learning}},
	publisher = {PMLR},
	author = {Batson, Joshua and Royer, Laurent},
	editor = {Chaudhuri, Kamalika and Salakhutdinov, Ruslan},
	year = {2019},
	pages = {524--533},
}

@inproceedings{denault_covariate-moderated_2025,
	title = {Covariate-moderated {Empirical} {Bayes} {Matrix} {Factorization}},
	url = {https://openreview.net/forum?id=OrmLtoFF60},
	doi = {10.48550/arXiv.2505.11639},
	language = {en},
	urldate = {2026-02-09},
	booktitle = {{NeurIPS}},
	author = {Denault, William R. P. and Tayeb, Karl and Carbonetto, Peter and Willwerscheid, Jason and Stephens, Matthew},
	month = oct,
	year = {2025},
}
